\documentclass[10pt, a4paper, twocolumn]{article}

\usepackage[utf8]{inputenc}
\usepackage[T1]{fontenc}
\usepackage[english]{babel}

\usepackage[top=2cm, bottom=2cm, left=1.5cm, right=1.5cm]{geometry}

\usepackage{mathpazo} 
\usepackage{microtype} 

\usepackage{graphicx}
\usepackage{xcolor}
\usepackage{float}
\usepackage{subcaption} 
\usepackage[font=small, labelfont=bf, labelsep=period]{caption}
\usepackage{amsmath, amssymb, amsfonts}

\usepackage{authblk} 
\usepackage{titlesec} 
\usepackage{abstract} 

\usepackage{booktabs} 
\usepackage{multirow}

\usepackage[colorlinks=true, linkcolor=blue, citecolor=blue, urlcolor=blue]{hyperref}

\titleformat{\section}{\large\bfseries}{\thesection}{1em}{}
\titleformat{\subsection}{\normalsize\bfseries}{\thesubsection}{1em}{}

\title{\textbf{\Large BertsWin: Resolving Topological Sparsity in 3D Masked Autoencoders via Component-Balanced Structural Optimization}}

\author[1]{Evgeny Alves Limarenko}
\author[1]{Anastasiia Studenikina}

\affil[1]{\small Moscow Institute of Physics and Technology, 141701 Dolgoprudny, Moscow Region, Russia}

\date{}

\begin{document}

\twocolumn[
  \begin{@twocolumnfalse}
    \maketitle
    \begin{abstract}
      \textit{\textbf{Abstract}---} 
      The application of self-supervised learning (SSL) and Vision Transformers (ViTs) approaches demonstrates promising results in the field of 2D medical imaging, but the use of these methods on 3D volumetric images is fraught with difficulties. Standard Masked Autoencoders (MAE), which are state-of-the-art solution for 2D, have a hard time capturing three-dimensional spatial relationships, especially when 75\% of tokens are discarded during pre-training. We propose BertsWin, a hybrid architecture combining full BERT-style token masking using Swin Transformer windows, to enhance spatial context learning in 3D during SSL pre-training. Unlike the classic MAE, which processes only visible areas, BertsWin introduces a complete 3D grid of tokens (masked and visible), preserving the spatial topology. And to smooth out the quadratic complexity of ViT, single-level local Swin windows are used. We introduce a structural priority loss function and evaluate the results of cone beam computed tomography of the temporomandibular joints. The subsequent assessment includes TMJ segmentation on 3D CT scans. We demonstrate that the BertsWin architecture, by maintaining a complete three-dimensional spatial topology, inherently accelerates semantic convergence by a factor of $5.8\times$ compared to standard ViT-MAE baselines. Furthermore, when coupled with our proposed GradientConductor optimizer, the full BertsWin framework achieves a 15-fold reduction in training epochs (44 vs 660) required to reach state-of-the-art reconstruction fidelity. Analysis reveals that BertsWin achieves this acceleration without the computational penalty typically associated with dense volumetric processing. At canonical input resolutions, the architecture maintains theoretical FLOP parity with sparse ViT baselines, resulting in a significant net reduction in total computational resources due to faster convergence.
      
      \vspace{0.5cm}
      \vspace{0.5cm}
    \end{abstract}
  \end{@twocolumnfalse}
]
\section{INTRODUCTION}

Temporomandibular joint (TMJ) disorders represent a heterogeneous group of pathologies affecting patients and manifesting as pain, functional impairment, and structural changes. The gold standard for diagnosis in this context is Cone-Beam Computed Tomography (CBCT), which provides a high-detail visualization of bone structures \cite{Yu2024}. Recent years have seen the active adoption of deep learning methods for diagnosing TMJ pathologies. Existing solutions, predominantly based on U-Net architectures or hybrid models combining the capabilities of Convolutional Neural Networks (CNNs) and Vision Transformers (ViTs), demonstrate high efficiency in segmenting anatomical structures, such as the mandibular condyle and mandibular canal \cite{Vinayahalingam2023, Gumussoy2025}.

Although 3D visualization of the TMJ allows for a comprehensive and detailed analysis, accounting for the spatial arrangement of structures, most deep learning research is conducted in 2D slices \cite{Manek2025}. This is primarily due to the high computational complexity of processing 3D data and the scarcity of labeled volumetric datasets, which require significant resources to create. A promising solution to this problem is the use of Self-Supervised Learning (SSL) algorithms, which allow the extraction of meaningful features from unlabeled images \cite{Huang2023}. Specifically, the Masked Autoencoders (MAE) approach, which reconstructs masked regions of an image, has become the state-of-the-art (SOTA) in 2D analysis \cite{He2022}.

However, the direct transfer of the MAE approach to 3D space faces several serious limitations. This is because aggressive masking, which involves removing up to 75\% of the information and is effective for planar images, leads to the destruction of the spatial structure and the loss of anatomical context in 3D \cite{Madan2025}. The random removal of voxels breaks the connections between adjacent structures, which is especially critical in medical diagnostics, where the integrity of the anatomical structures studied are paramount \cite{Wald2025}. An additional obstacle is the high computational complexity of processing full 3D token grids, which limits the deployment of algorithms under conditions of hardware resource constraints \cite{Takahashi2024}.

In this situation, a fundamental redesign of encoder architectures is warranted. Specifically, the creation of new encoders capable of ''understanding'' and preserving complex spatial topology and structural information, rather than merely reconstructing pixels. Based on this, we propose the BertsWin architecture, which implements a two-stage encoder. In the first stage, a 3D CNN stem processes only the visible subset of patches. These are then scattered into a full 3D token grid to restore the original spatial framework. In the second stage, a single-level Swin Transformer is applied, which, through its window mechanism, focuses on reconstructing the local context while considering the 3D neighborhood. To enhance the quality of reconstruction, we have integrated a specialized additional loss function, ''Mean Variance Correlation Loss'' (MVC Loss), which decomposes the error into components of brightness, contrast, and structure, allowing the learning focus to shift towards the balanced reconstruction of all diagnostically significant image parameters.

\section{RELATED WORK}

ViTs have revolutionized 2D medical imaging, surpassing traditional CNNs by modeling global context and capturing long-range dependencies in clinical tasks \cite{Halder2024}. CNNs are limited by local receptive fields, whereas ViTs use self-attention to process image patches and capture global spatial relationships \cite{Li2024}. Hybrid CNN-ViT models have shown particular effectiveness in distinguishing closely spaced structures in MRI images, such as intervertebral discs and TMJ components \cite{Yoon2024}.

Although there is currently no specific literature on the successful application of ViT to 3D CBCT images for TMJ pathologies, Jiang's study suggests that the 3D Swin Transformer architecture outperformed the standard 3D U-Net in brain tumor segmentation, indicating the potential of this approach for CBCT imaging \cite{Jiang2022}. Their application to 3D CBCT images for TMJ pathology detection faces two primary obstacles such as computational complexity when processing volumetric data and the scarcity of large annotated datasets specific to TMJ disorders.

SSL addresses the annotated data shortage by extracting meaningful features from unlabeled medical images through their inherent structure and patterns \cite{Huang2023}. Recent advances include ToothMCL, a multimodal pre-training system that constructs volumetric tooth segmentation from CBCT scans while preventing information leakage between adjacent slices \cite{Son2025}. However, SSL pre-training demands substantial computational resources and large unlabeled datasets (typically thousands to tens of thousands of images), with requirements increasing exponentially when transitioning from 2D to 3D imaging \cite{Huang2023}.

MAE offers a computationally efficient SSL approach by processing only visible image regions -- typically masking 75\% of image patches and training encoders to reconstruct missing areas \cite{He2022}. This aggressive masking exploits image redundancy and substantially reduces computational cost \cite{Zhang2023}. Nevertheless, applying MAE to volumetric medical data introduces critical challenges. Furthermore, the asymmetric design of standard MAE, which assumes spatial locality holds in two dimensions, does not adequately account for the fact that volumetrically distant voxels may be anatomically adjacent in 3D space, potentially compromising learned representations \cite{Wald2025, Madan2025}.

The evolution from convolutional encoders to transformer-based and hybrid architectures reflects the medical imaging community's progression toward capturing both local spatial patterns and global anatomical dependencies. A critical but often overlooked consideration is the alignment between encoder architecture, SSL pre-training objectives, and downstream clinical tasks. MAE-based encoders prioritize spatial coherence and local structure preservation due to their pixel-level reconstruction focus, whereas contrastive learning encoders (e.g., SimCLR, MoCo) benefit from learning augmentation-invariant representations. Recent work demonstrates that jointly optimizing pre-training objectives and architecture selection -- rather than treating them as independent choices -- improves generalization to downstream segmentation, classification, and detection tasks \cite{He2022, Wald2025}.

Optimization algorithm selection further influences encoder pre-training effectiveness. Large batch SSL pre-training benefits from optimizers like LARS (Layer-wise Adaptive Rate Scaling), which stabilizes learning across massive batch sizes by adaptively scaling layer-wise learning rates based on parameter-to-gradient norm ratios. Empirical evidence shows that LARS-optimized SSL models significantly outperform SSL in low-data regimes. The limitations of LARS include instability in the training process during early iterations without Warm-Up and a predisposition to Sharp Minimizers due to an excessively high layer-wise normalization rate, which leads to an uncontrolled increase in the scaled gradient \cite{Do2024}. Another notable optimizer is LION (EvoLved Sign Momentum), a memory-efficient optimizer designed to reduce GPU memory requirements by approximately 50\% compared to AdamW while maintaining competitive performance, making it suitable for downstream fine-tuning with limited labeled data. However, the original LION optimizer can be prone to issues such as explosion or gradient attenuation, making it unreliable for medical applications \cite{Rong2025}. Despite theoretical promise, this two-stage optimization paradigm -- LARS for pre-training, LION for fine-tuning -- remains largely unexplored in medical imaging, and current studies rarely specify which optimizers were employed, obscuring their contribution to model performance.

Loss function choice fundamentally determines which features encoders learn to extract during SSL pre-training. Mean squared error (MSE -- L2 loss) dominates due to computational efficiency and mathematical simplicity, but assumes errors are normally distributed, independent, and uniformly important across pixels -- assumptions often violated in medical imaging \cite{Terven2025, Sara2019}. More critically, MSE-optimized models tend to produce overly smoothed reconstructions that sacrifice clinically significant fine details, as MSE treats intensity shifts and structural information loss equivalently despite their perceptually different impacts on diagnostic quality \cite{Sara2019}.

An alternative is the Structural Similarity Index (SSIM) better aligns with human visual perception by incorporating brightness and contrast masking phenomena \cite{Sun2022, DeJonge2023}. SSIM decomposes into brightness, contrast, and structural components, enabling identification of specific reconstruction failure modes relevant to clinical interpretation \cite{Renieblas2017}. SSIM is formulated as:
\begin{equation*}
\text{SSIM}(I, \hat{I}) = \frac{(2\mu_i \hat{\mu}_i + C_1)(2\sigma_{i\hat{i}} + C_2)}{(\mu_i^2 + \hat{\mu}_i^2 + C_1)(\sigma_i^2 + \hat{\sigma}_i^2 + C_2)}
\end{equation*}
where $C_1$ and $C_2$ are the stabilization constants, $\sigma_{i\hat{i}}$ represents the covariance between the images. 

However, SSIM's multiplicative formulation causes disproportionate value decreases when any component fails substantially, and SSIM-only optimization can introduce artifacts in homogeneous regions while experiencing gradient conflicts \cite{Renieblas2017}.
Despite strong mathematical correlation (Pearson $r = 0.967\text{--}0.994$), MSE and SSIM exhibit complementary properties that justify their combined use \cite{Wang2004}:
\begin{equation*}
\min \text{MSE}(x,y) \iff \max |\text{SSIM}(x,y)| \iff \max |r_{xy}|
\end{equation*}
MSE preserves absolute intensity values, which are critical for applications such as radiation therapy planning, while SSIM maintains spatial structure and edge definition essential for diagnostic imaging. Combined optimization improves reconstruction quality compared to either metric individually across diverse medical imaging applications \cite{Sun2022}.

One possible way to overcome this limitation is to attempt to decompose MSE into interpretable components -- brightness error, contrast error, and structural error -- through variance decomposition, enabling targeted control over each aspect during optimization \cite{Hafeez2024, Elharrouss2025}. This Mean Variance Correlation decomposition provides more informative gradients but remains underutilized, with most studies employing MSE without differentiating individual component contributions. 

Thus, several serious gaps in the field of medical 3D visualization have been identified. First, adoption of specialized optimizers such as LARS and LION in medical imaging remains limited despite demonstrated effectiveness in general computer vision, primarily due to conservative institutional practices using Adam and a lack of medical imaging specific benchmarks. Second, the co-design of encoder architecture, SSL objectives, loss functions, and optimization algorithms remains underexplored, with most approaches treating these as independent rather than coupled system components. Third, domain-specific encoder architectures and loss weightings tailored to particular clinical tasks such as TMJ.

\section{MATERIALS AND METHODS}

\subsection{Dataset Construction and Preprocessing}
\textit{Data Acquisition.} The dataset comprised 2 TB of CBCT scans acquired over two years using a Vatech Green X18 (PHT-75CHS) device. All participants provided written informed consent prior to data collection. Personally identifiable information was systematically removed from DICOM headers, with each scan assigned a unique anonymous identifier in compliance with established ethical standards and data protection regulations.

\textit{ROI Extraction Pipeline.} From the 2 TB collection, scans with maximum resolution (0.2 mm isotropic spacing, $900^3$ voxels) were selected for analysis. An automated pipeline performed: bone segmentation based on Hounsfield Unit (HU) thresholding to separate upper and lower jaw structures; TMJ localization via extreme occipital point detection on the mandible; extraction of $400\times400\times400$ voxel ROIs corresponding to an $80\times80\times80$ mm physical volume centered on each TMJ; and storage in HDF5 format. This yielded a unified 3D joint dataset.

\textit{Data Partitioning.} Patient-level partitioning, rather than scan-level partitioning, was implemented to prevent information leakage between training and validation sets. The final data distribution consisted of 6,550 training joints and 1,156 validation joints. Z-normalization statistics, including HU clipping percentiles, mean and standard deviation, were computed exclusively from the training data.

\subsection{BertsWin Architecture}
\begin{figure}[t] 
    \centering
    \includegraphics[width=\columnwidth]{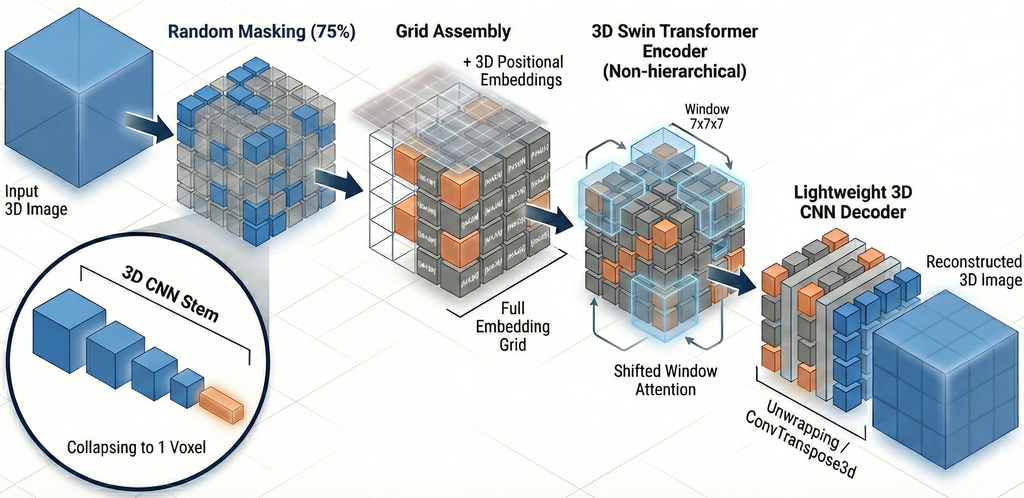} 
    \caption{The architecture of BertsWin.}
    \label{fig:bertswin_arch}
\end{figure}

\textit{Overview.} We introduce BertsWin, a hybrid framework designed for volumetric MAE. Departing from the standard asymmetric MAE paradigm -- which processes only visible patches -- our architecture preserves the complete spatial grid by incorporating learnable mask tokens prior to the encoding stage. This design allows the model to leverage Shifted Window Attention mechanisms on the full volumetric structure. The architecture is composed of three distinct modules: a multi-stage 3D CNN stem for dense local feature extraction, an isotropic, non-hierarchical Swin Transformer encoder, and a lightweight convolutional decoder for voxel-wise reconstruction, as illustrated in Fig.~\ref{fig:bertswin_arch}.

\begin{figure*}[t]
    \centering
    \begin{subfigure}[b]{0.48\textwidth}
        \centering
        \includegraphics[width=\linewidth]{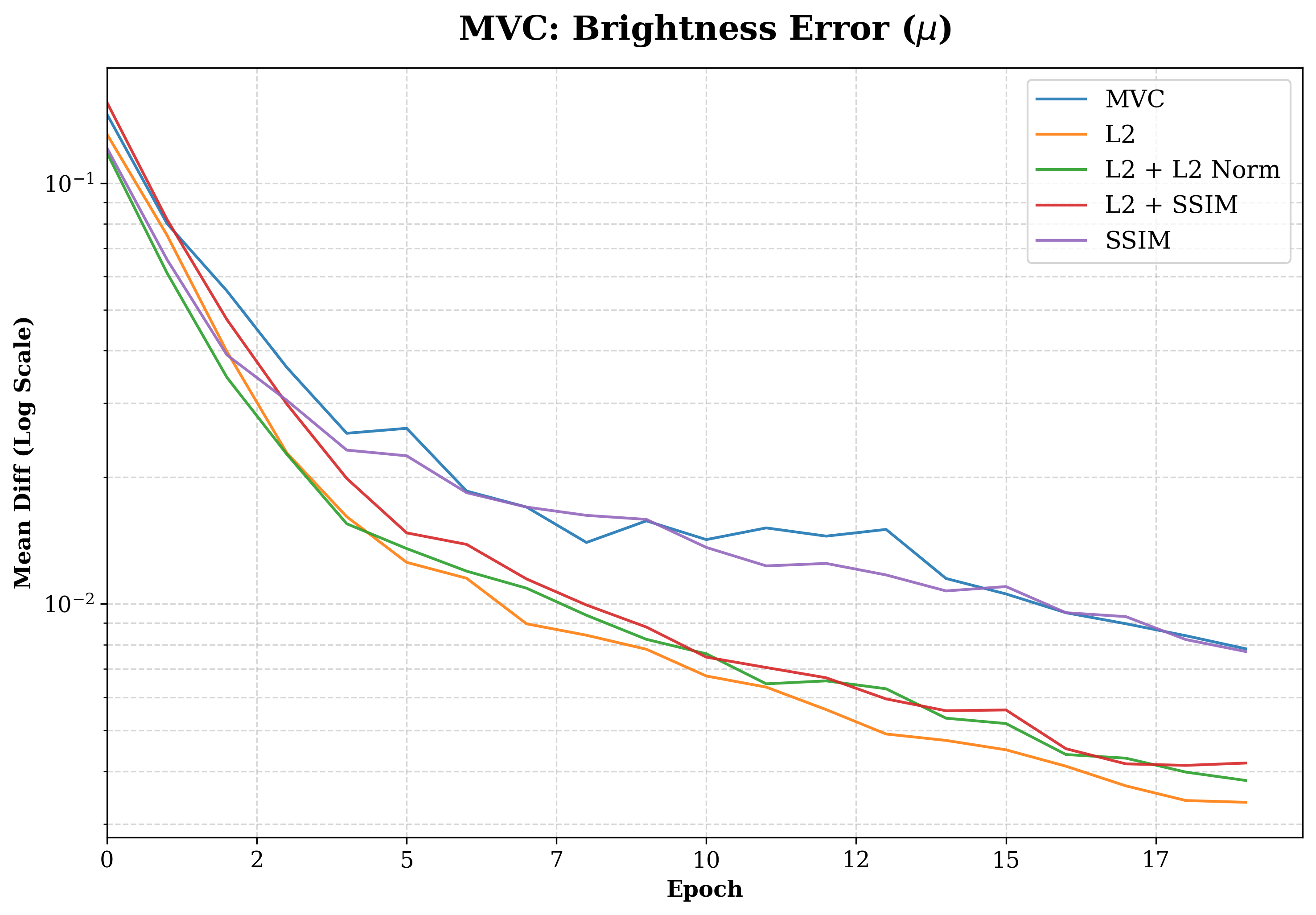}
        \caption{Brightness Error}
        \label{fig:mvc_brightness}
    \end{subfigure}
    \hfill 
    \begin{subfigure}[b]{0.48\textwidth}
        \centering
        \includegraphics[width=\linewidth]{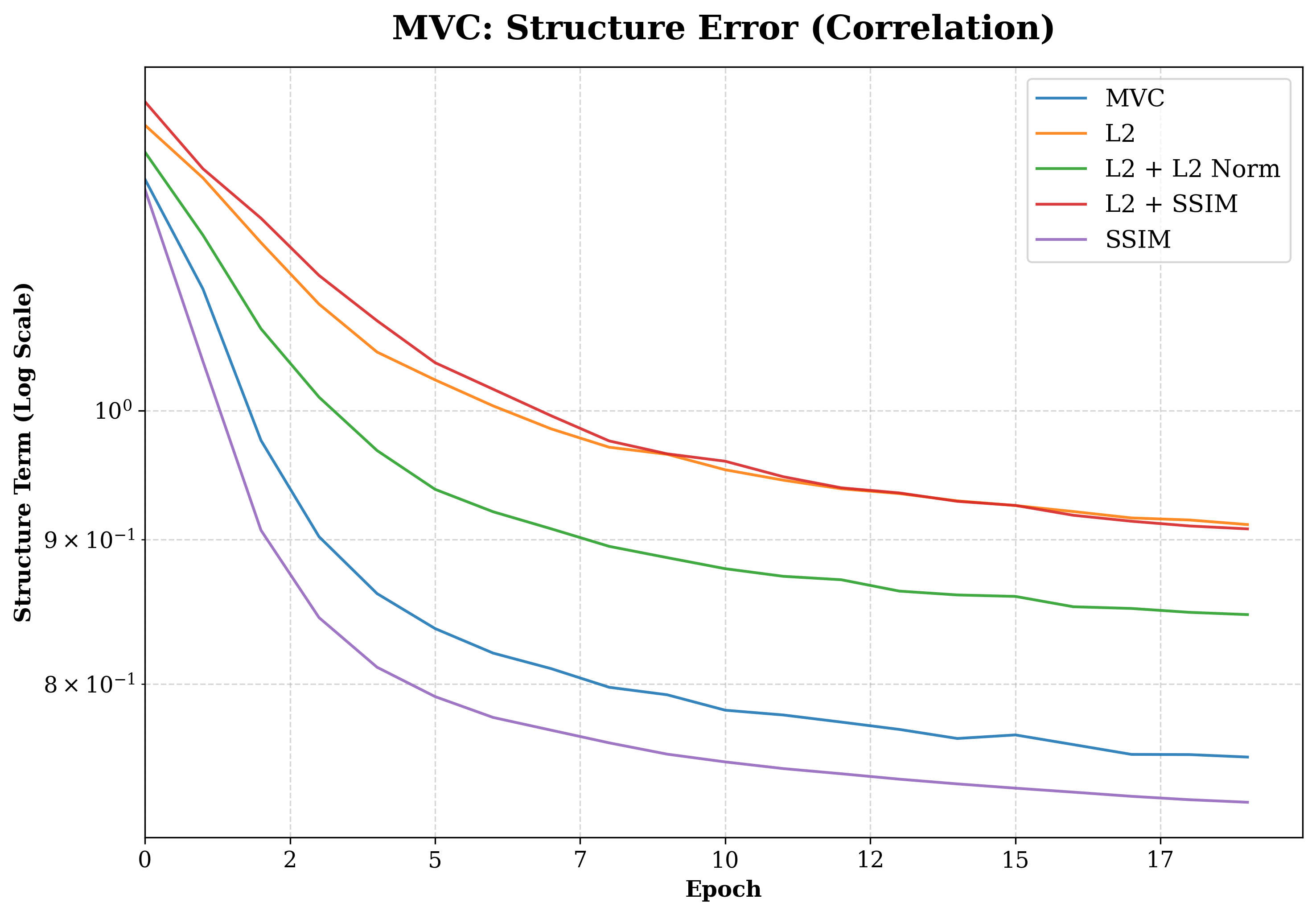}
        \caption{Structure Error}
        \label{fig:mvc_structure}
    \end{subfigure}
    
    \caption{Loss Function Components and Rationale for Multi-Component Variance (MVC) Regularization. Our gradient analysis during the MAE training process on head and neck CT images confirmed an inherent conflict within the $L2$-loss. The optimization process consistently prioritizes the minimization of the mean intensity component (Brightness Error).}
    \label{fig:mvc_analysis}
\end{figure*}

\textit{Baseline: MONAI MAE-ViT.} The baseline implementation utilizes the standard MAE-ViT architecture from the Medical Open Network for Artificial Intelligence (MONAI) framework \cite{Cardoso2022}. This architecture divides a $224^3$ voxel input volume into 2,744 patches (each $16^3$ voxels), randomly masks 75\% of the patches, and processes only the remaining 686 visible patches through a ViT-Base encoder consisting of 12 transformer layers before reconstruction.
\paragraph{Proposed Components.}
\textbf{Hybrid Patch Embedding.} A compressive CNN stem consisting of four 3D convolutional blocks with stride 2 reduces each $16^3$ voxel patch to $1^3$ voxel representation before projection into the transformer embedding space.

\textbf{Full Token Grid.} Rather than discarding masked patches, a complete token grid of dimensions $(B, 2744, C)$ is constructed, where $B$ represents batch size and $C$ represents channel dimension. Visible patch embeddings are scattered to their original spatial positions, while remaining positions are populated with learnable MASK tokens. Positional embeddings are subsequently added to preserve spatial structure throughout the network.

\textbf{Single-Scale Swin Encoder.} Twelve Swin Transformer blocks with 12 attention heads and $7\times7\times7$ window sizes process the complete $14^3$ token grid without hierarchical downsampling. This single-resolution processing enables intensive local context analysis that is better suited for reconstruction tasks compared to hierarchical feature extraction.

\textbf{Lightweight Decoder.} A CNN-based upsampling decoder comprising three transposed 3D convolutional layers with strides of 4, 2, and 2 reconstructs the full $224^3$ voxel volume from the encoder output.

\textbf{Design Rationale.} The patch size of $16^3$ voxels, corresponding to $3.2^3$ mm in physical space, was determined based on TMJ anatomical considerations and was found to be optimal for encoding cortical bone structures and articular surfaces. The input volume size of $224^3$ voxels ensures complete joint coverage, including surrounding soft tissues critical for comprehensive diagnostic assessment.

\section{Training Configuration}

\subsection{Analysis of Loss Function Components and Rationale for Multi-Component Variance Regularization}

The conventional MSE ($L2$-loss) is a standard objective in image reconstruction. However, its application in SSL MAE training on medical data, such as Computed Tomography, presents optimization challenges due to the high dynamic range and domain-specific feature priorities. We justify the departure from $L2$ by conducting an analysis of its underlying components.

\subsubsection{The Component Conflict in L2}
The MSE can be mathematically decomposed into fundamental image components related to brightness, contrast, and structure, a principle utilized in the Multi-Component Variance (MVC) decomposition. This is expressed as:
\begin{equation*}
\small
\setlength{\medmuskip}{1mu}
\text{MSE}(X,Y) = (\mu_X - \mu_Y)^2 + (\sigma_X - \sigma_Y)^2 + 2\sigma_X \sigma_Y (1 - \rho(X,Y))
\end{equation*}
Where $X$ and $Y$ are the predicted and target patches, and $\mu$, $\sigma$, and $\rho$ denote the mean (brightness), standard deviation (contrast) and Pearson correlation coefficient (structure), respectively.

Analysis of gradient dynamics in head and neck CT images identified an inherent conflict within the L2-loss, which is corroborated by the suboptimal structural convergence shown in (Fig.~\ref{fig:mvc_analysis}). The optimization process consistently prioritizes the minimization of the mean intensity component, $(\mu_X - \mu_Y)^2$. In high-dynamic-range CT data, this bias results in the model focusing primarily on restoring high-magnitude pixel regions (i.e., bone structures) as this yields the largest instantaneous reduction in the total squared error. This dominance of the brightness component causes the model trajectory to converge to a local minimum that is suboptimal for structural fidelity.

\subsubsection{Implementation of Weighted MVC Loss}
To counteract this component dominance and achieve a balanced feature representation, we employed a weighted additive loss function based on the MVC principle:
\begin{equation}
L_{\text{MVC}} = w_{\text{Br}} \cdot L_{\text{Br}} + w_{\text{Cntr}} \cdot L_{\text{Cntr}} + w_{\text{Str}} \cdot L_{\text{Str}}
\end{equation}
Where $L_{\text{Br}}$, $L_{\text{Cntr}}$, and $L_{\text{Str}}$ represent the loss components calculated across image patches. $L_{\text{Str}}$ is the $L2$ on normalized (zero-mean, unit-variance) patches (often termed $L2\text{Norm}$).

The necessity of this approach is validated by the convergence profiles of the respective models. As illustrated by the convergence graphs (Fig.~\ref{fig:mvc_analysis} (left) for the Br metric and Fig.~\ref{fig:mvc_analysis} (right) for the Str metric), models trained solely with $L2$ exhibit rapid saturation in the Br metric but demonstrate significantly hindered progress in minimizing the Str error. The MVC approach, however, successfully modifies the optimization path, enabling concurrent and balanced progress across all components.

A dedicated ablation study was performed to determine the optimal weighting coefficients for the primary task. The configuration yielding the most informative visual reconstruction -- characterized by balanced restoration of both high-density bone structures and low-density soft tissues -- was determined to be: $w_{\text{Br}}=0.3$, $w_{\text{Cntr}}=0.2$, and $w_{\text{Str}}=0.5$. This configuration explicitly regularizes the MAE towards high-fidelity structural restoration, which is critical for the downstream tasks of medical image analysis.

\subsubsection{Anatomy-Aware Physically Guided Loss Formulation}
While the MVC decomposition addresses the structural optimization conflict, the application of MAE to CBCT of the TMJ introduces a domain-specific challenge: intensity imbalance. The high pixel intensity of osseous structures dominates the global loss calculation, potentially suppressing gradients derived from low-contrast soft tissues (e.g., the articular disc and pterygoid muscles), which are critical for comprehensive diagnosis.

To mitigate this, we propose a physically guided loss, \textit{PhysLoss}, which extends the MVC framework by applying anatomical masking and region-specific prioritization. This formulation computes the additive MVC loss ($L_{\text{MVC}}$) separately across three semantically defined regions:
\begin{equation}
\begin{aligned}
L_{\text{phy}} &= \lambda_{\text{global}}L_{\text{MVC}}(\Omega) + \lambda_{\text{soft}}L_{\text{MVC}}(M_{\text{soft}}) + \\
&\quad \lambda_{\text{surf}}L_{\text{MVC}}(M_{\text{surf}})
\end{aligned}
\end{equation}
Where $\Omega$ represents the full patch domain. The masking strategies and weighting coefficients $\lambda$ are determined based on distinct clinical and physical priors.

\paragraph{Anatomical Mask Generation.} The masks are generated dynamically from the target patches to guide the reconstruction focus:

\textbf{Soft Tissue Mask ($M_{\text{soft}}$).} Defined by intensity thresholds derived from the dataset-specific HU histogram analysis. We target the range $[-300, 300]$ HU (Z-normalized), which encompasses the signal distribution of the TMJ soft tissues while excluding air and dense cortical bone.

\textbf{Bone Surface Shell ($M_{\text{surf}}$).} Diagnosis of TMJ osteoarthritis relies heavily on the integrity of the condylar surface (e.g., detecting erosions). To capture this high-frequency boundary, we generate a morphological shell around the bone segmentation. We apply asymmetric morphological operations: an inner erosion (kernel size $k=2$) and an outer dilation ($k=4$). This asymmetry, established in consultation with domain experts, ensures the mask captures both the inner cortical thickness and the immediate external transition zone, reducing partial volume effects at the boundary.

\paragraph{Signal-Equalized Weighting Strategy.} The weights are assigned using a signal-equalization strategy rather than arbitrary tuning.

\noindent $\bullet$ $\lambda_{\text{soft}}=0.5$: Soft tissues in CBCT possess a low Contrast-to-Noise Ratio. To prevent feature collapse in these regions, we assign the highest weight, effectively amplifying their contribution to the gradient and counteracting the dominance of high-intensity bone pixels.

\noindent $\bullet$ $\lambda_{\text{surf}}=0.2$: The bone surface inherently possesses strong gradients due to the sharp intensity transition. While it does not require the same amplification as soft tissue, it requires explicit separation from the bulky trabecular bone to ensure the model focuses on high-frequency surface topology rather than internal volume density.

\noindent $\bullet$ $\lambda_{\text{global}}=0.3$: The residual weight maintains global spatial coherence and regularizes the reconstruction of background elements.

Finally, to maximize local feature extraction within these masks, the MVC components are computed on $8^3$ sub-patches within each $16^3$ encoding patch, ensuring that the statistics (mean, variance, correlation) are calculated with sufficient local granularity.

\subsubsection{GradientConductor Optimizer}
For all SSL experiments, we used an effective batch size of 192 ($N=192$). We found that for our relatively homogeneous dataset, this batch size provides a fairly stable gradient. This stability allows us to abandon the resource-intensive AdamW (which stores the 1st and 2nd moments) in favor of more efficient optimizers. We have developed GradientConductor (GCond), a module that implements the LION/LARS optimizer with bias correction \cite{Limarenko2025}. The algorithm combines the sign-based update rule of LION with the layer-wise adaptive scaling of LARS and the bias correction mechanism of Adam. The update rule at step $t$ for parameter $p$ with gradient $g_t$ is defined as:

\begin{equation}
\begin{aligned}
m_{t} &= \beta_{1}m_{t-1} + (1-\beta_{1})g_{t} \\
\hat{m}_{t} &= m_{t} / (1-\beta_{1}^{t}) \quad \text{(Bias Correction)} \\
\lambda_{t} &= \min\left(\frac{\|p_{t-1}\|}{\|\hat{m}_{t}\|+\epsilon}, \lambda_{\text{clip}}\right) \quad \text{(Trust Ratio)} \\
p_{t} &= p_{t-1} - \eta \cdot \gamma \cdot \lambda_{t} \cdot \text{sign}(\hat{m}_{t})
\end{aligned}
\end{equation}

Where $\eta \cdot \gamma$ represents the effective learning rate coefficient. In our experiments, we utilize a fused coefficient set to $1.5 \times 10^{-5}$, which balances the aggressive sign-based updates of LION with the adaptive scaling of LARS. Unlike the standard LION optimizer, we explicitly incorporate bias correction ($m_t$) to stabilize the magnitude of the trust ratio $\lambda_t$ during the initial ``warm-up'' phase.

This approach offers distinct advantages for volumetric 3D training. First, it acts as a memory-efficient alternative to AdamW, reducing optimizer state VRAM usage by $\approx 50\%$ (storing only the first moment). Second, by integrating LARS-style layer-wise scaling with the sign-based update of LION, it normalizes gradient magnitudes across the varying depths of the hybrid CNN-Transformer architecture, preventing the ``vanishing gradient'' issues often observed in deep volumetric encoders. Crucially, while bias correction does not alter the update direction (due to the sign operation), it fundamentally rescales the momentum norm ($\|m_t\|$), ensuring a stable trust ratio ($\lambda_t$) estimation even in early training iterations when the moving average is initialized at zero.

\subsection{Evaluation Protocol}
\textbf{Feature Extraction.} For BertsWin, features were extracted using adaptive average pooling on the $14^3$ encoder feature map. For the ViT baseline, global average pooling was applied across all patch tokens, excluding the CLS token, as this token is not optimized for MAE-based feature aggregation.

\subsection{Implementation Details and Computational Considerations}
\textbf{Weight Transfer Protocol for MAE Baseline.} 
A structural incompatibility was identified within the MONAI framework between the self-supervised MAE ViT architecture used for pre-training and the standard ViT architecture required for downstream tasks. Direct weight loading is obstructed by discrepancies in layer definitions. To enable transfer learning, we implemented a specific weight mapping protocol: the target ViT classifier was instantiated with \texttt{pos\_embed\_type='sincos'}, followed by a manual alignment of the LayerNorm weights from the final MAE encoder block (\texttt{blocks[-1]}) to the standard architecture's normalization layer (\texttt{ViT.norm}). This procedure ensured correct weight initialization prior to feature extraction.

\subsection{Computational Complexity and Throughput Analysis}

We performed a theoretical complexity analysis comparing the proposed full-grid BertsWin against the sparse (25\% visible) MAE-ViT baseline. Results are summarized in Table~\ref{tab:complexity}.

\textbf{Standard Resolution ($224^3$, P16).} At the nominal resolution, BertsWin Base demonstrates computational parity with the MONAI ViT Base (223.84 vs 228.29 GFLOPs). The linear complexity ($O(N)$) of the windowed attention mechanism effectively offsets the four-fold increase in token count required for full-grid processing, negating the computational advantage of sparse masking in this regime.

\textbf{High Resolution ($512^3$, P16).} With fine-grained patches, the baseline ViT exhibits quadratic complexity growth ($O(N^2)$), reaching $\sim11.0$ TFLOPs. In contrast, BertsWin maintains linear scalability ($\sim2.7$ TFLOPs), yielding a $4.1\times$ reduction in computational cost per epoch.

\textbf{Coarse Patch Trade-off ($512^3$, P32).} A complexity inversion is observed at larger patch sizes. The sparse baseline leverages the reduced token count to lower global attention overhead (658 GFLOPs), whereas BertsWin retains a higher fixed cost (1272 GFLOPs). This identifies a specific trade-off: BertsWin is computationally optimal for fine-grained (P16) volumetric analysis, whereas sparse masking retains efficiency advantages in coarse-token regimes.

\begin{table}[h]
    \centering
    \caption{Computational complexity comparison in GFLOPs. Complexity is estimated as $2\times$MACs with a fixed masking ratio of 75\% for the encoder.}
    \label{tab:complexity}
    \small 
    \setlength{\tabcolsep}{3pt} 
    \begin{tabular}{lrrrr}
        \toprule
        \textbf{Model} & \textbf{Stem} & \textbf{Encoder} & \textbf{Decoder} & \textbf{Total} \\
        \midrule
        \multicolumn{5}{l}{\textit{Configuration: $224^3$ (P16)}} \\
        BertsWin Base & 81.7 & 125.2 & 16.9 & \textbf{223.8} \\
        BertsWin Small & 80.9 & 33.5 & 9.5 & \textbf{123.9} \\
        MONAI ViT Base & 17.3 & 134.1 & 76.9 & 228.3 \\
        \midrule
        \multicolumn{5}{l}{\textit{Configuration: $512^3$ (P16)}} \\
        BertsWin Base & 976.0 & 1495.2 & 201.9 & \textbf{2673.1} \\
        BertsWin Small & 966.4 & 399.7 & 113.8 & \textbf{1479.9} \\
        MONAI ViT Base & 206.2 & 3866.2 & 6963.1 & 11035.5 \\
        \midrule
        \multicolumn{5}{l}{\textit{Configuration: $512^3$ (P32)}} \\
        BertsWin Base & 1026.8 & 186.9 & 59.1 & 1272.7 \\
        BertsWin Small & 1024.4 & 50.0 & 48.1 & 1122.4 \\
        MONAI ViT Base & 206.2 & 212.9 & 239.1 & \textbf{658.1} \\
        \bottomrule
    \end{tabular}
\end{table}

The standard ViT baseline (MONAI) exhibits quadratic complexity growth ($O(N^2)$) at high resolutions with small patches ($512^3$, P16), reaching $\sim11$ TFLOPs. In contrast, BertsWin demonstrates linear scalability ($O(N)$) due to windowed attention, while maintaining high-fidelity feature extraction in the CNN stem.

The application of the GCond optimizer further reduces this total computational budget by achieving convergence in just 44 epochs ($15.0\times$ speedup). Thus, the proposed full-grid approach is more computationally efficient in terms of total GPU-hours required to obtain a viable model. While the GCond architecture supports multi-objective gradient projection, in this study it was deployed in 'Fused Mode', treating the weighted PhysLoss as a singular scalar objective to focus purely on the efficiency of the LION-LARS optimization trajectory.

Regarding system throughput, 3D volumetric SSL imposes unique constraints distinct from 2D vision. The training bottleneck shifts from compute (FLOPs) to data ingestion bandwidth. Our preliminary benchmarks on high-end monolithic accelerators (NVIDIA H200) revealed significant GPU idle time due to PCIe bus saturation when feeding large batches. To mitigate this, we employed a distributed data-parallel setup with four NVIDIA RTX 5090 GPUs. This configuration parallelizes the I/O streams, effectively bypassing the single-bus bottleneck.

Unlike standard ViT architectures where computational cost explodes quadratically at high resolutions ($512^3$, P16), BertsWin maintains a linear trajectory ($O(N)$), resulting in a $\sim4.1\times$ reduction in FLOPs (2.67 vs 11.04 TFLOPs). While sparse baselines retain an efficiency edge in coarse-patch regimes (P32) due to lower token counts, BertsWin remains the only computationally viable architecture for full-grid fine-grained volumetric analysis at this scale without relying on aggressive masking.

\subsection{Embedding Quality Analysis}

\textbf{Feature Distribution Analysis.} Primary embedding analysis included visualization of feature activation distributions at the encoder output ($N = 12,288$ features from the reference ``Core Cohort'' batch) using kernel density estimation (KDE) to assess feature utilization and potential feature collapse.

\textbf{Probing Tasks for Semantic Assessment.} For comprehensive semantic evaluation, a suite of probing tasks was applied to a dedicated validation sample using cosine similarity as the evaluation metric. The following properties were assessed: invariance to spatial transformations and HU value augmentations; inter-patient separability; intra-patient similarity between canonical left and right joints; and anatomical symmetry understanding (left versus horizontally flipped right joints).

\textbf{Statistical Analysis.} Statistical analysis of cosine similarity distributions between models employed descriptive statistics (median, interquartile range) and 95\% confidence intervals (CI) obtained through bootstrap resampling. The Mann-Whitney U-test ($p < 0.001$) with 95\% CI using bias-corrected and accelerated (BCa) bootstrap ($n = 9,999$ iterations) was applied for each comparison. Additionally, to assess statistical significance of differences between intra-patient similarity (L-R) and symmetry similarity (L-R-flipped) within each model, the paired nonparametric Wilcoxon signed-rank test was employed.

\subsection{Downstream Segmentation Task}
To provide an unbiased assessment of the intrinsic semantic capacity of the pre-trained encoders, we employed a linear probing protocol. This approach involved freezing the encoder weights and training only a lightweight linear projection head ($1\times1$ convolution), thereby isolating feature quality from decoder learning capability. Domain-specific models (Ours, MONAI MAE/ViT) were evaluated at their native $224^3$ resolution. Conversely, to ensure a fair comparison, the generalist SAM Med3D baseline \cite{Wang2025} -- pre-trained at $128^3$ -- was evaluated by downsampling input volumes to $128^3$ via trilinear interpolation, maintaining fidelity to its official weights. 

The Ground Truth (GT) for this task was curated semi-automatically to eliminate inter-observer variability. Initial masks were generated on the full-resolution source CBCT volumes using the robust DentalSegmentator \cite{Dot2024}. These predictions were then refined via a cascade of heuristic filters -- evaluating component connectivity, Center of Mass (CoM) anatomical validity, and contralateral symmetry -- resulting in a high-fidelity cohort of 1,152 joints with a strict patient-level split to prevent data leakage.

\subsection{Sensitivity Analysis: Anatomical Chirality and Orientation}
To verify that the learned representations encode distinct morphological geometry rather than spatially invariant textures, we conducted a paired Wilcoxon signed-rank test comparing Intra-Patient similarity (Real Right vs Contralateral Left, mirrored to valid orientation) against a Symmetry Check (Real Right vs Artificially Mirrored Right). It should be borne in mind that the training distribution consisted exclusively of right-sided joints (with left-sided samples mirrored during preprocessing). Consequently, the model learned a latent manifold specific to right-sided anatomical chirality. In the Intra-Patient test, the contralateral left joint is mirrored, effectively transforming it into a valid right-sided anatomical instance. This aligns the sample with the learned manifold. In the Symmetry Check, the right joint is mirrored, artificially creating a left-sided anatomical instance. From the encoder's perspective, this represents an out-of-distribution sample with reversed anatomical gradients (e.g., medial-lateral inversion).
\begin{figure*}[t]
    \centering
    \begin{subfigure}[b]{0.32\textwidth}
        \centering
        \includegraphics[width=\linewidth]{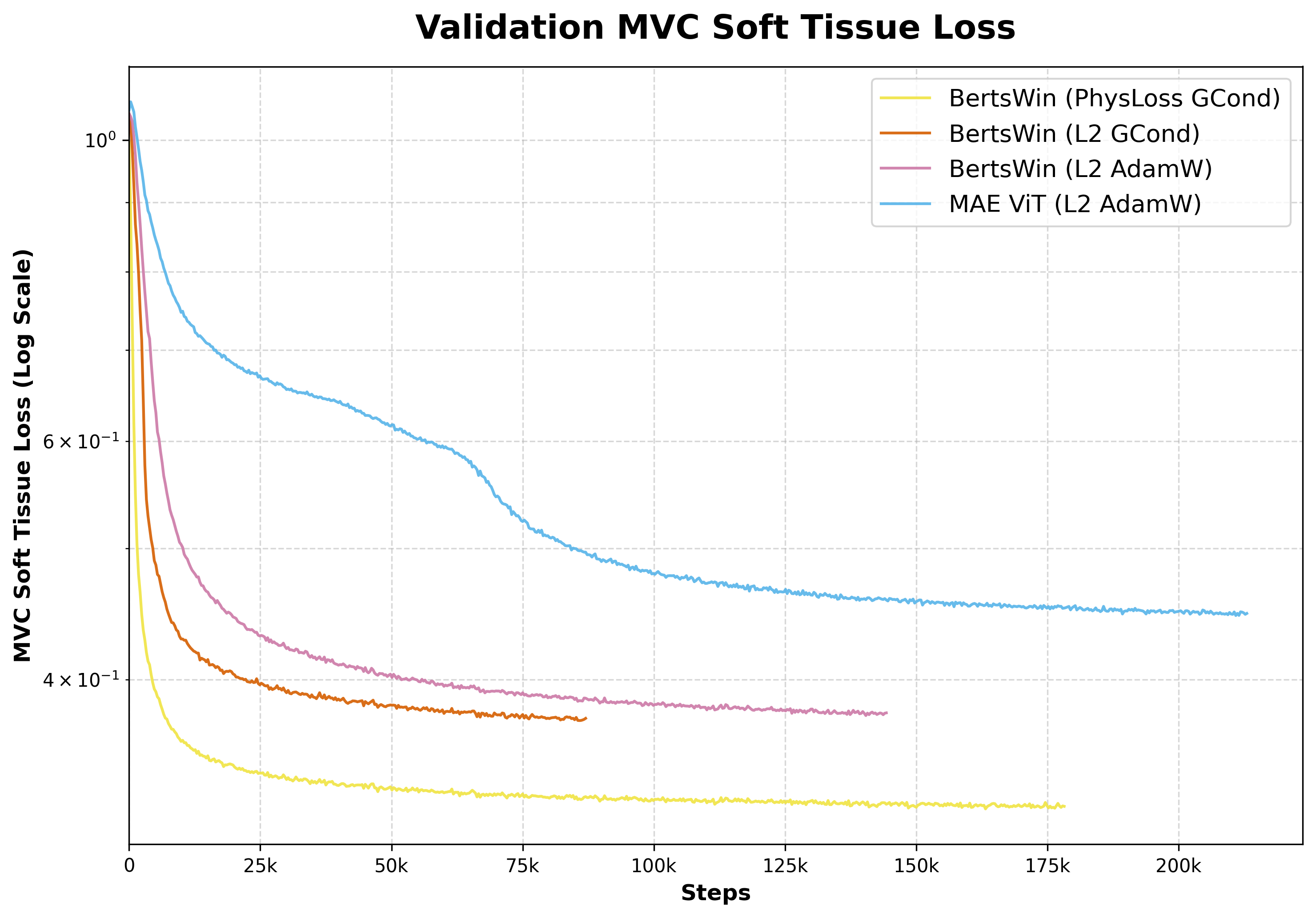}
        \caption{MVC: Soft Tissue}
        \label{fig:conv_soft}
    \end{subfigure}
    \begin{subfigure}[b]{0.32\textwidth}
        \centering
        \includegraphics[width=\linewidth]{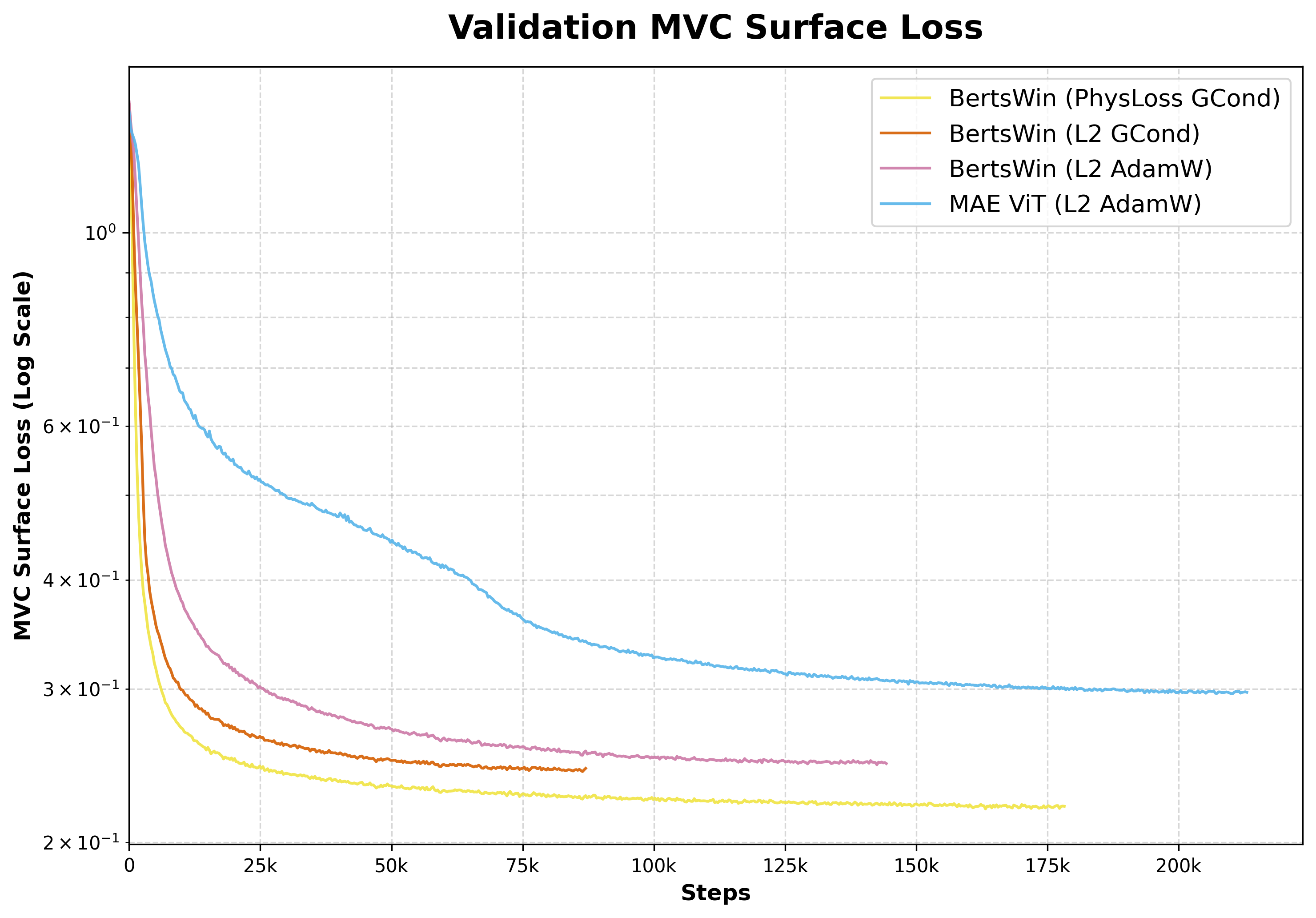}
        \caption{MVC: Surface}
        \label{fig:conv_surface}
    \end{subfigure}
    \begin{subfigure}[b]{0.32\textwidth}
        \centering
        \includegraphics[width=\linewidth]{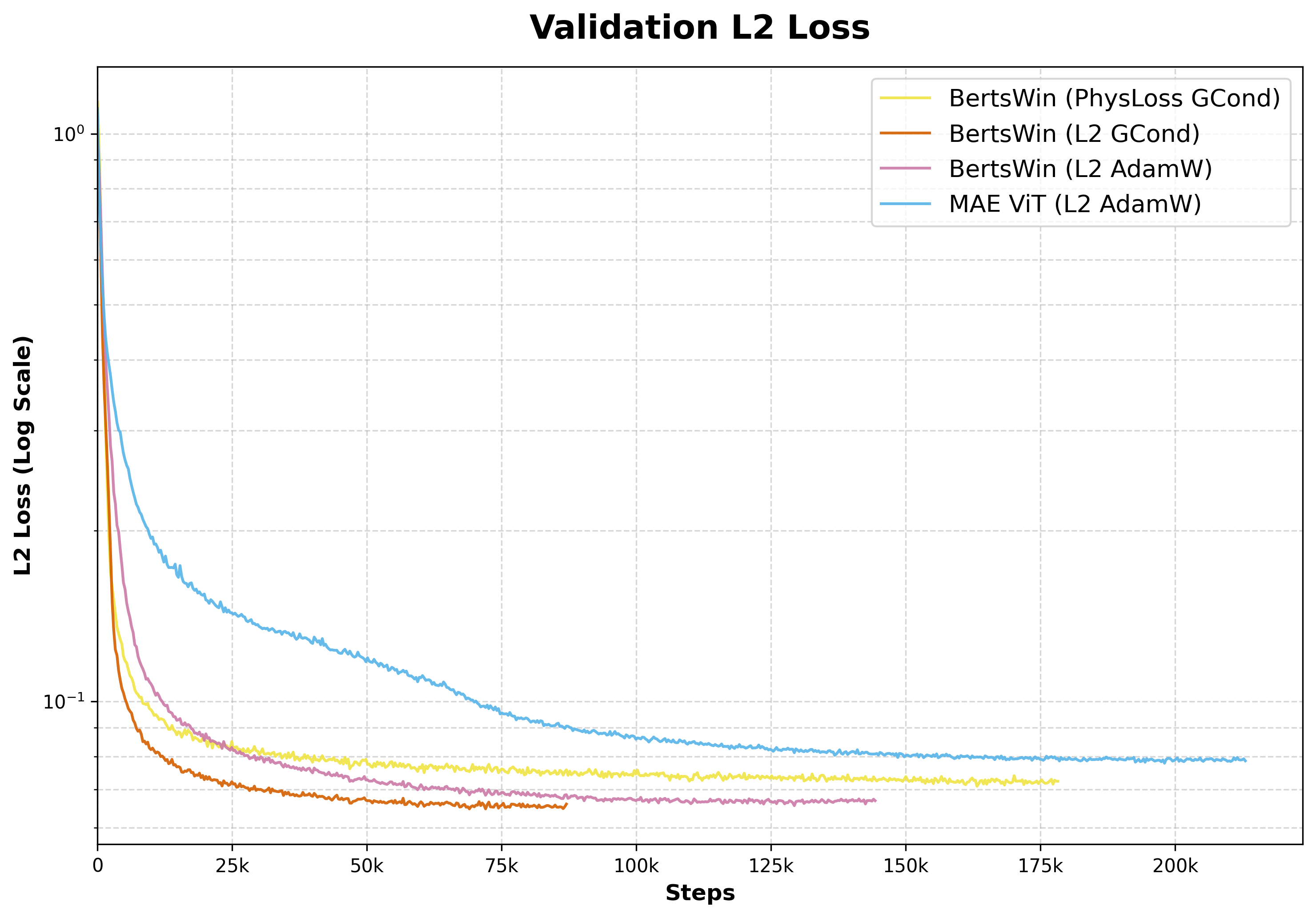}
        \caption{L2 Loss}
        \label{fig:conv_l2}
    \end{subfigure}
    
    \caption{Validation Convergence Dynamics and Optimization Efficiency. Comparison of BertsWin models with various loss functions and optimizers (PhysLoss GCond, L2 GCond, L2 AdamW) and the basic MAE ViT model (L2 AdamW). (a) Dynamics of the MVC metric for soft tissues; (b) MVC metric for surfaces; (c) Dynamics of L2 losses on the validation set. The x-axis shows the number of learning steps, and the y-axis shows the metric values on a logarithmic scale.}
    \label{fig:convergence}
\end{figure*}

\section{RESULTS}

\subsection{Training dynamics and convergence analysis}

A comparative analysis of training dynamics and convergence revealed a fundamental disparity in optimization efficiency between the proposed BertsWin architecture and the baseline MAE ViT (Fig.~\ref{fig:convergence}, Table~\ref{tab:convergence}). The proposed BertsWin (PhysLoss) achieves the lowest validation error for both Soft Tissue and Surface structures (lowest $1-\text{MVC}$), significantly outperforming the L2-optimized baselines. The introduction of the GCond optimizer further accelerates the BertsWin convergence to 44 epochs compared to MAE ViT (epoch 660). While our optimizer could accelerate the baseline MAE ViT as well, our objective is to compare the proposed BertsWin ecosystem against the established community standard (MONAI implementation with default AdamW guidelines). Even isolating the architectural BertsWin demonstrates superior learning efficiency compared to MAE ViT.

\begin{table}[h]
    \centering
    \caption{Validation Convergence Dynamics and Optimization Efficiency.}
    \label{tab:convergence}
    \small
    \setlength{\tabcolsep}{6pt} 
    \begin{tabular}{lccr} 
        \toprule
        \textbf{Model} & \textbf{Best L2} & \textbf{Epoch} & \textbf{Speedup} \\
        \midrule
        BertsWin (PhysLoss GCond) & 0.0708 & 152 & $4.3\times$ \\
        BertsWin (L2 GCond) & \textbf{0.0646} & \textbf{44} & $\mathbf{15.0\times}$ \\
        BertsWin (L2 AdamW) & 0.0655 & 114 & $5.8\times$ \\
        MAE ViT (L2 AdamW) & 0.0778 & 660 & $1.0\times$ \\
        \bottomrule
    \end{tabular}
\end{table}

Crucially, this acceleration is achieved without incurring a computational penalty per iteration. As shown in the complexity analysis, processing the full token grid at $224^3$ incurs approximately the same FLOP cost as the sparse baseline (223 vs 228 GFLOPs). The efficiency of local Swin windows neutralizes the cost of processing masked tokens, meaning the 15-fold reduction in training epochs translates directly to net computational savings. Furthermore, the learning trajectories exhibit distinct characteristics. The MAE ViT baseline displays a period of instability (a ``hump'') between steps 25k and 75k, particularly evident in soft tissue metrics, while BertsWin demonstrates a monotonic reduction in error from the earliest iterations.

\subsection{Comparative analysis of latent space representations}

\begin{figure}[H]
    \centering
    \includegraphics[width=\columnwidth]{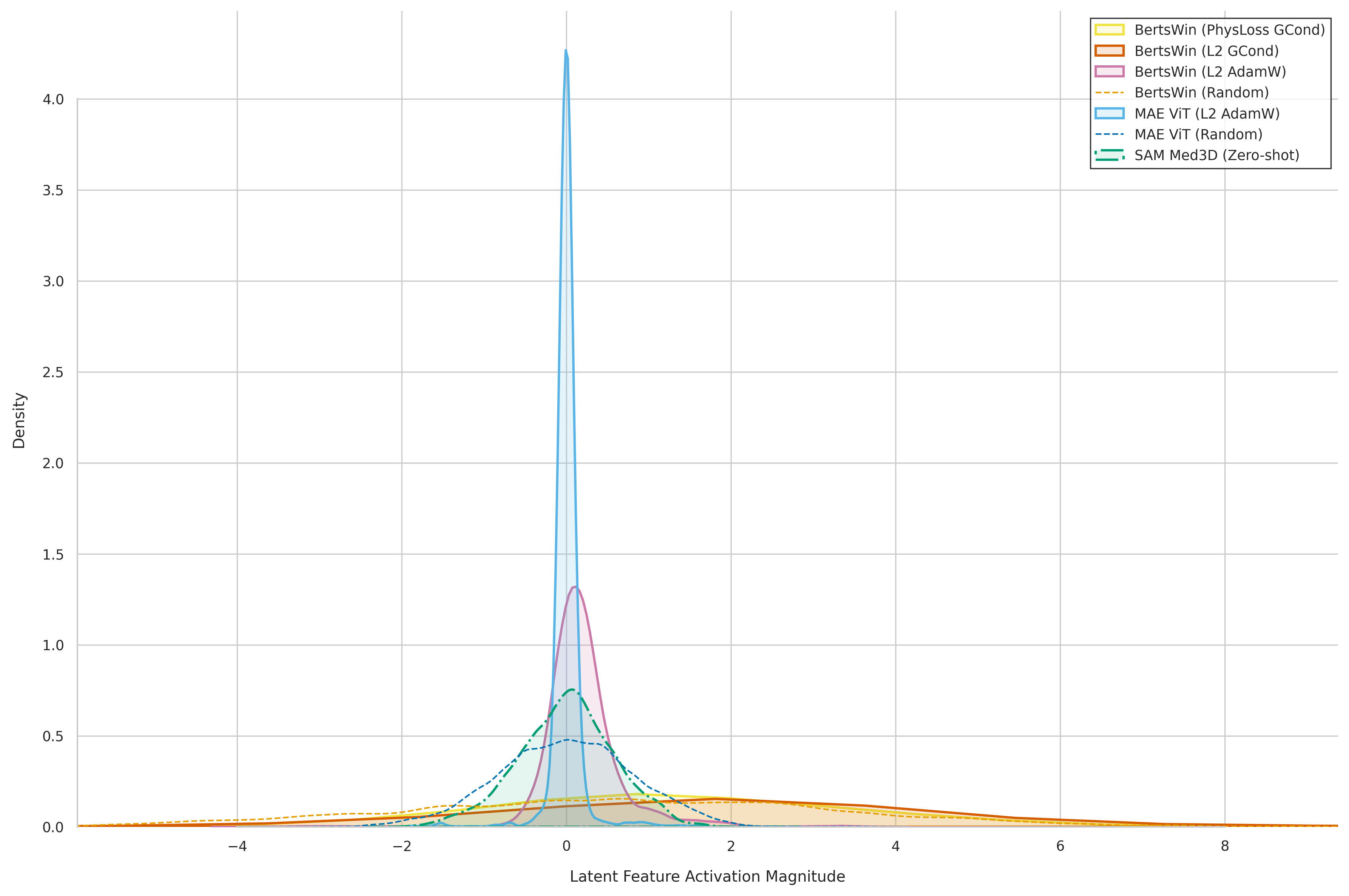}
    \caption{Comparative Density of Latent Representations. Visual analysis of feature magnitude distributions. Random networks (dashed) display high-entropy Gaussian-like profiles. Solid curves correspond to trained models, while dashed curves denote randomly initialized networks.}
    \label{fig:latent_density}
\end{figure}

A comparison of the density of latent space representations (Fig.~\ref{fig:latent_density}) showed that MAE ViT baseline training (blue curve) results in a narrow distribution centered near zero, indicative of feature suppression or collapse. In contrast, the PhysLoss-guided framework (yellow) maintains a broad dynamic range with a structured shift in activation density, demonstrating that the model learns a non-trivial, semantically rich manifold without succumbing to the representation collapse typical of voxel-wise MSE optimization.

To evaluate the semantic richness and structural integrity of the learned representations, we analyzed feature vectors $\mathbf{z} \in \mathbb{R}^d$ extracted by the frozen encoders from whole volumes of the validation dataset.

\paragraph{Mitigating Representation Anisotropy.} Preliminary spectral analysis revealed that the feature spaces of all trained transformers (both ViT and Swin-based) exhibit the ``cone effect'' (representation degeneration), where embeddings collapse into a narrow cone in the vector space. Under these conditions, standard Euclidean metrics or Principal Component Analysis (PCA) are dominated by the vector magnitude rather than direction.

To mitigate this anisotropy, we employed Pearson correlation as the similarity metric. This is mathematically equivalent to cosine similarity on centered and normalized embeddings:
\begin{equation}
\text{sim}(\mathbf{u}, \mathbf{v}) = \frac{(\mathbf{u} - \bar{\mathbf{u}}) \cdot (\mathbf{v} - \bar{\mathbf{v}})}{\|\mathbf{u} - \bar{\mathbf{u}}\|_2 \|\mathbf{v} - \bar{\mathbf{v}}\|_2}
\end{equation}
This approach neutralizes global intensity shifts caused by acquisition device variability, allowing us to compare the relative activation patterns of neurons -- effectively the ``anatomical fingerprint.''

\paragraph{Spectral Topology and Intrinsic Dimensionality.}
We analyzed the intrinsic dimensionality using the Effective Rank ($R_{\text{eff}}$) of the feature covariance matrix (Table~\ref{tab:latent_topology}). This analysis revealed a fundamental spectral divergence between MSE-optimized models (including baselines) and structure-aware encoders. The MAE ViT baseline exhibits a high effective rank ($R_{\text{eff}} \approx 23.36$). However, coupled with its low geometric invariance ($0.549$). In contrast, the proposed BertsWin (PhysLoss GCond) compresses the representation to a low effective rank ($R_{\text{eff}} \approx 4.01$). Crucially, this aligns closely with the zero-shot SAM Med3D foundation model ($R_{\text{eff}} \approx 4.54$), while significantly diverging from the noise-dominated baseline.

\begin{table}[h]
    \centering
    \caption{Quantitative Analysis of Latent Space Topology.}
    \label{tab:latent_topology}
    \small
    \setlength{\tabcolsep}{3pt}
    \begin{tabular}{lcrcl}
        \toprule
        \textbf{Model Configuration} & \textbf{Dim} & \textbf{$R_{\text{eff}}$} & \textbf{Geo. Inv.} & \textbf{Intra-Pat.} \\
        \midrule
        MAE ViT (Random) & 768 & 3.25 & 0.823 & 0.718* \\
        BertsWin (Random) & 768 & 3.07 & 0.796 & 0.793* \\
        SAM Med3D (Zero-shot) & 384 & 4.54 & 0.780 & 0.662 \\
        \midrule
        MAE ViT (L2 + AdamW) & 768 & 23.36 & 0.549 & 0.329 \\
        BertsWin (L2 + AdamW) & 768 & 9.52 & 0.619 & 0.525 \\
        BertsWin (L2 + GCond) & 768 & 10.99 & 0.628 & 0.472 \\
        BertsWin (MVC + GCond) & 768 & 6.43 & 0.739 & 0.635 \\
        \textbf{BertsWin (PhysLoss)} & \textbf{768} & \textbf{4.01} & \textbf{0.831} & \textbf{0.791} \\
        \bottomrule
    \end{tabular}
    \vspace{0.2cm}
    \parbox{\columnwidth}{\footnotesize \textit{Note:} Comparison of spectral properties, robustness, and discriminative power. ``Random'' baselines establish the isometry floor. High similarity scores in random projections (*) are an artifact of the Johnson-Lindenstrauss isometry property. Unlike optimized models, these baselines fail to minimize Inter-Patient correlation, lacking semantic discriminative power.}
\end{table}

\begin{figure*}[t]
    \centering
    \includegraphics[width=0.8\textwidth]{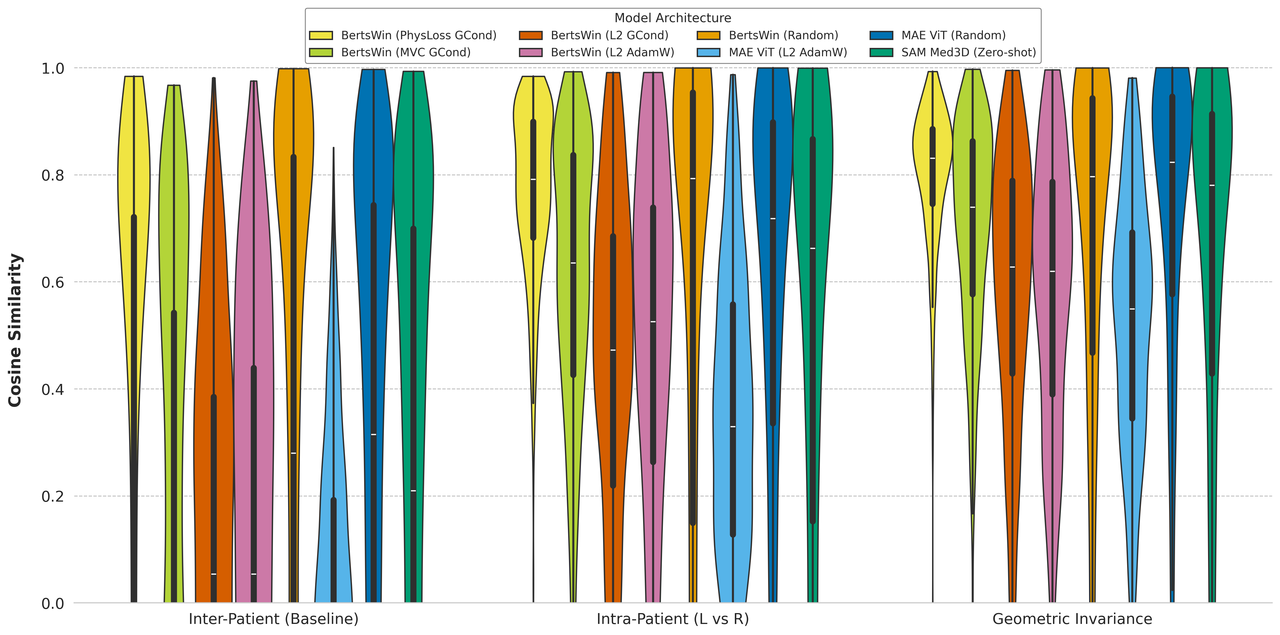}
    \caption{Feature Space Discriminability and Invariance Analysis. Violin plots visualize the density distribution of cosine similarity scores across three evaluation protocols. The internal box plots represent the median and interquartile range (IQR). Inter-Patient serves as the baseline noise floor (lower is better), while Intra-Patient and Geometric Invariance represent signal preservation and robustness (higher is better).}
    \label{fig:violins}
\end{figure*}

All models trained with Mean Squared Error (MSE), including MAE ViT, BertsWin (AdamW), and BertsWin (GCond), converged to a suboptimal local minimum characterized by low geometric invariance (Median $0.549\text{--}0.628$) and poor intra-patient similarity (Median $< 0.53$). Transitioning to structural objectives (MVC and PhysLoss) resulted in a pronounced quantitative shift in representation quality. The median Intra-Patient similarity increased significantly from $0.472$ (L2) to $0.791$ (PhysLoss).

The initialized BertsWin (Random) exhibits a deceptively competitive mean geometric invariance ($0.645$), statistical dispersion analysis reveals its lack of semantic stability, whereas the IQR for the random network is $0.476$ -- substantially wider than the optimized BertsWin (PhysLoss) ($\text{IQR} = 0.140$). Furthermore, the random baseline fails to segregate patients, showing a high positive Inter-Patient similarity (median = $0.280$), whereas the optimized model effectively orthogonalizes distinct patients (median = $-0.120$).

\subsection{Sensitivity Analysis: Anatomical Chirality and Orientation}

A paired Wilcoxon signed-rank test comparing Intra-Patient similarity (Real Right vs Contralateral Left, mirrored to valid orientation) against symmetry checks (Real Right vs Artificially Mirrored Right) showed that the proposed BertsWin (PhysLoss GCond) demonstrates a statistically significant preference for the Intra-Patient pairs ($p < 0.001$), with a median similarity differential of $-0.039$ favoring the contralateral joint over the synthetic reflection. Moreover, it demonstrates the largest discriminative margin (separation) between inter-patient noise and intra-patient signal (Fig.~\ref{fig:violins}), significantly outperforming L2-based baselines which exhibit overlapping distributions.

\begin{figure}[H]
    \centering
    \includegraphics[width=\columnwidth]{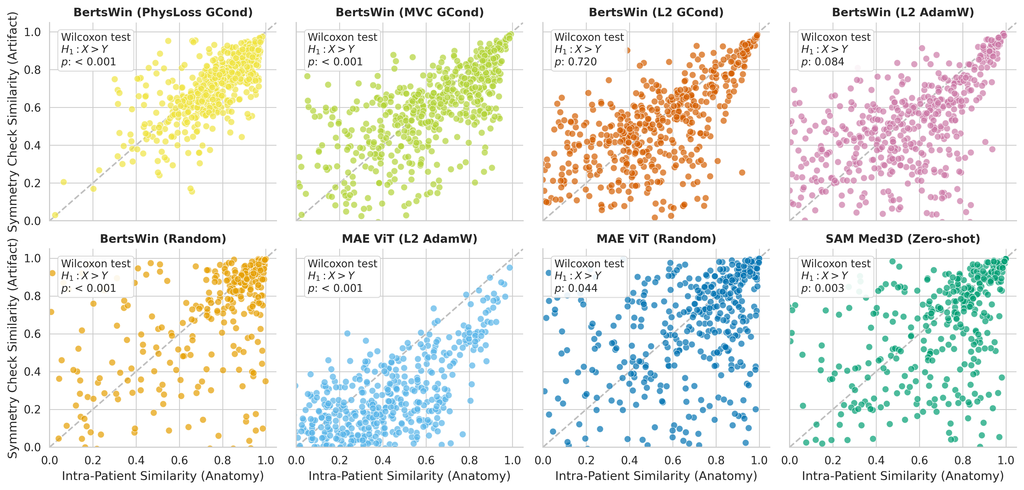}
    \caption{Paired Sensitivity and Semantic Confidence Analysis. Scatter plots comparing feature similarity for real contralateral pairs (Intra-Patient, x-axis) versus artificially mirrored pairs (Symmetry, y-axis).}
    \label{fig:scatter}
\end{figure}

In the analysis of paired sensitivity and semantic confidence (Fig.~\ref{fig:scatter}), the points below the dashed diagonal ($y = x$) indicate that the model prioritizes valid anatomical chirality over synthetic mirroring artifacts. Crucially, the cluster location acts as a proxy for semantic confidence: while baselines like MAE ViT fall below the diagonal but remain trapped in the low-similarity noise regime (bottom-left, $< 0.4$), the proposed BertsWin (PhysLoss) demonstrates a dense concentration in the high-similarity regime (top-right, $> 0.7$).

\subsection{Reconstruction fidelity and loss function priors}

\begin{figure}[H]
    \centering
    \includegraphics[width=0.45\textwidth]{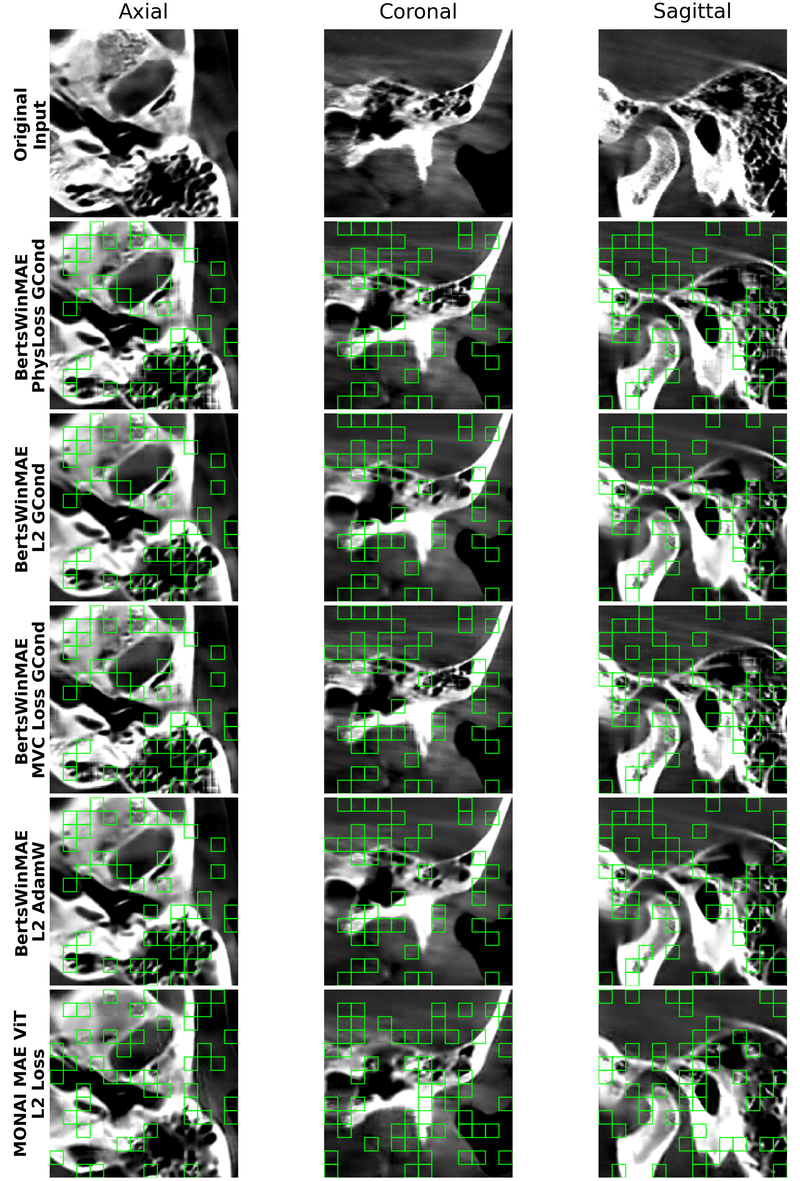}
    \caption{Qualitative assessment of volumetric inpainting performance across different architectures. Note: The visualization displays orthographic projections (Axial, Coronal, Sagittal) of a representative test sample. Green outlines indicate the visible patches (25\% context) provided to the model input; the remaining regions represent the masked anatomy reconstructed by the network.}
    \label{fig:2d_slice}
\end{figure}

\paragraph{2D Slice Analysis: Internal Structure and Soft Tissue.}
The analysis of orthogonal projections of the inference result for a sample from the validation dataset (Fig.~\ref{fig:2d_slice}) allows for the assessment of internal structural recovery. While all evaluated models demonstrate high global anatomical consistency, successfully recovering the overall geometry of the mandible, distinct textural differences are observed. Models optimized via L2-loss (both the baseline MAE ViT and BertsWin-L2) exhibit a tendency toward spectral smoothing, especially noticeable in areas of soft tissues and fine structural details. In contrast, approaches utilizing MVC-based objectives (PhysLoss GCond, MVC Loss GCond) yield sharper reconstructions with superior preservation of the internal trabecular bone structure and cortical boundaries.

\begin{figure}[H]
    \centering
    \includegraphics[width=0.39\textwidth]{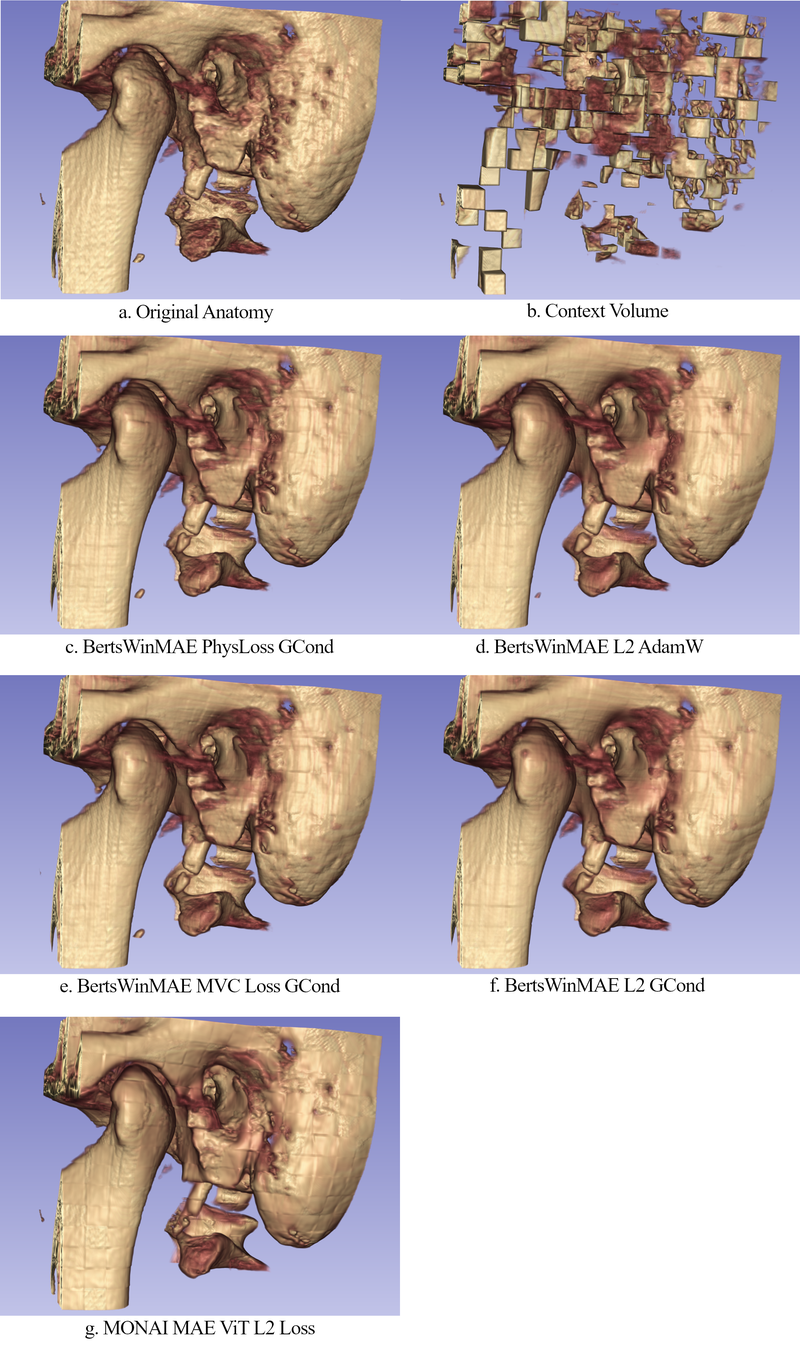}
    \caption{Comparative visualization of volumetric reconstruction of the complete TMJ. Note: Isosurface renderings of the complete TMJ complex. (a) Ground Truth: Original anatomical volume. (b) Masked Input: Visualization of the 25\% random visible patches provided to the encoder.}
    \label{fig:3d_iso}
\end{figure}

\paragraph{3D Methodology and Spatial Integrity.}
To evaluate spatial topology, isosurface rendering of osseous structures was performed (Fig.~\ref{fig:3d_iso}) using the bone rendering preset in 3D Slicer. For experimental rigor, 25\% of the visible patches were integrated into the reconstructed volume in their original form (contextual injection). This step is necessary to isolate the generative performance on masked regions 75\%, as the standard MAE objective does not strictly guarantee the identity reconstruction of visible tokens.

Comparative visualization of volumetric reconstruction of the complete TMJ showed that BertsWin (PhysLoss) reconstructs the most physically plausible surface texture, preserving trabecular heterogeneity. BertsWin (L2 AdamW) demonstrates consistent global geometry with characteristic L2-induced surface smoothing. BertsWin (MVC Loss) enhances structural definition but exhibits minor residual transitions at patch boundaries due to asynchronous loss convergence. BertsWin (L2 GCond) achieves the highest spatial coherence, forming a monolithic, seamless volume, though lacking high-frequency micro-texture. MONAI MAE ViT: Fails to resolve global spatial context, resulting in severe fragmentation and blocking artifacts.

\subsection{Downstream Segmentation Task}
\begin{table*}[t]
    \centering
    \caption{Comparative segmentation performance evaluation using linear probing. Note: All models were evaluated using a frozen encoder and a trained linear projection head ($1\times1$ convolution). Values represent mean $\pm$ standard error.}
    \label{tab:segmentation}
    \small
    \begin{tabular}{lrrrrrr}
        \toprule
        \multirow{2}{*}{\textbf{Model}} & \multicolumn{3}{c}{\textbf{Mandible}} & \multicolumn{3}{c}{\textbf{Skull}} \\
        \cmidrule(lr){2-4} \cmidrule(lr){5-7}
        & \textbf{Dice} & \textbf{HD95} (mm) & \textbf{ASSD} (mm) & \textbf{Dice} & \textbf{HD95} (mm) & \textbf{ASSD} (mm) \\
        \midrule
        BertsWinMAE L2 GCond & 0.777 $\pm$ 0.004 & 3.92 $\pm$ 0.15 & \textbf{1.17 $\pm$ 0.04} & 0.878 $\pm$ 0.003 & 3.47 $\pm$ 0.48 & 0.80 $\pm$ 0.05 \\
        BertsWinMAE L2 AdamW & 0.774 $\pm$ 0.001 & 4.53 $\pm$ 0.15 & 1.53 $\pm$ 0.09 & 0.869 $\pm$ 0.001 & 3.25 $\pm$ 0.25 & 0.76 $\pm$ 0.03 \\
        BertsWinMAE PhysLoss GCond & 0.760 $\pm$ 0.002 & 4.84 $\pm$ 0.16 & 1.61 $\pm$ 0.11 & 0.827 $\pm$ 0.001 & 10.04 $\pm$ 0.16 & 1.82 $\pm$ 0.02 \\
        BertsWinMAE MVC Loss GCond & 0.749 $\pm$ 0.002 & 4.69 $\pm$ 0.15 & 1.34 $\pm$ 0.01 & 0.803 $\pm$ 0.004 & 12.40 $\pm$ 0.54 & 2.82 $\pm$ 0.11 \\
        BertsWinMAE Random & 0.540 $\pm$ 0.005 & 8.97 $\pm$ 0.13 & 2.89 $\pm$ 0.05 & 0.555 $\pm$ 0.002 & 9.47 $\pm$ 0.23 & 3.09 $\pm$ 0.10 \\
        MONAI MAE ViT Random & 0.688 $\pm$ 0.003 & 13.00 $\pm$ 0.06 & 2.64 $\pm$ 0.05 & 0.272 $\pm$ 0.022 & 26.53 $\pm$ 0.27 & 9.56 $\pm$ 0.28 \\
        MONAI MAE ViT L2 Loss & \textbf{0.785 $\pm$ 0.001} & \textbf{3.74 $\pm$ 0.13} & 1.23 $\pm$ 0.04 & \textbf{0.903 $\pm$ 0.001} & \textbf{2.53 $\pm$ 0.03} & \textbf{0.49 $\pm$ 0.01} \\
        SAM Med3D & 0.626 $\pm$ 0.000 & 6.87 $\pm$ 0.13 & 2.00 $\pm$ 0.06 & 0.691 $\pm$ 0.002 & 10.43 $\pm$ 0.27 & 2.45 $\pm$ 0.04 \\
        \bottomrule
    \end{tabular}
\end{table*}
To contextualize the quality of the learned features, we included the image encoder of the foundation model SAM Med3D as a reference point. It is crucial to clarify the intent of this comparison: we do not evaluate SAM Med3D's full segmentation capability, which relies on prompt engineering and a specialized mask decoder. Rather, we employ its ViT-Base encoder as a fixed feature extractor operating at its native $128^3$ resolution. While this restricts spatial granularity, it strictly maintains the model's valid operating distribution, avoiding artifacts associated with out-of-distribution inputs. Consequently, SAM Med3D serves here not as a direct segmentation baseline, but as a proxy for quantifying the spectral domain gap. The performance disparity highlights the orthogonality between generalist medical features (trained on soft-tissue MRI/CT) and the specific mineralized density distributions inherent to dental CBCT, validating the necessity of domain-specific representation learning.

The segmentation performance evaluated via linear probing is summarized in Table~\ref{tab:segmentation}. The domain-specific baseline, MONAI MAE ViT (L2 Loss), achieved the highest volumetric overlap metrics ($\text{DSC}_{\text{Mandible}} = 0.785$, $\text{DSC}_{\text{Skull}} = 0.903$). The proposed BertsWinMAE (L2 GCond) achieved competitive parity with this baseline ($\text{DSC}_{\text{Mandible}} = 0.777$, $\text{DSC}_{\text{Skull}} = 0.878$), confirming that the hybrid Swin-based architecture successfully learns spatial topology comparable to standard ViT models within the linear evaluation regime. In contrast, the foundation model SAM Med3D demonstrated a significant performance gap ($\text{DSC}_{\text{Mandible}} = 0.626$) compared to domain-specific encoders. Regarding architectural biases in untrained networks, MONAI MAE (Random) outperformed BertsWinMAE (Random) ($\text{DSC}_{\text{Mandible}}$ 0.688 vs 0.540), reflecting the difference in signal propagation between local convolutional embeddings and deep 3D CNN stems.

A distinct divergence was observed in boundary precision metrics based on the pre-training objective. Models optimized with L2 loss (both MONAI and BertsWin) consistently minimized geometric error ($\text{HD95} < 4.0$ mm), aligning with the density-based Ground Truth generated by DentalSegmentator. Conversely, models trained with structural objectives (PhysLoss, MVC) exhibited higher Hausdorff distances ($\text{HD95} > 10.0$ mm for the Skull) despite maintaining high volumetric overlap ($\text{DSC} > 0.74$). This metric discrepancy reflects a deviation between the topologically closed predictions of structural models and the densitometric, threshold-based definitions of the reference masks.

\begin{figure}[H]
    \centering
    \includegraphics[width=\columnwidth]{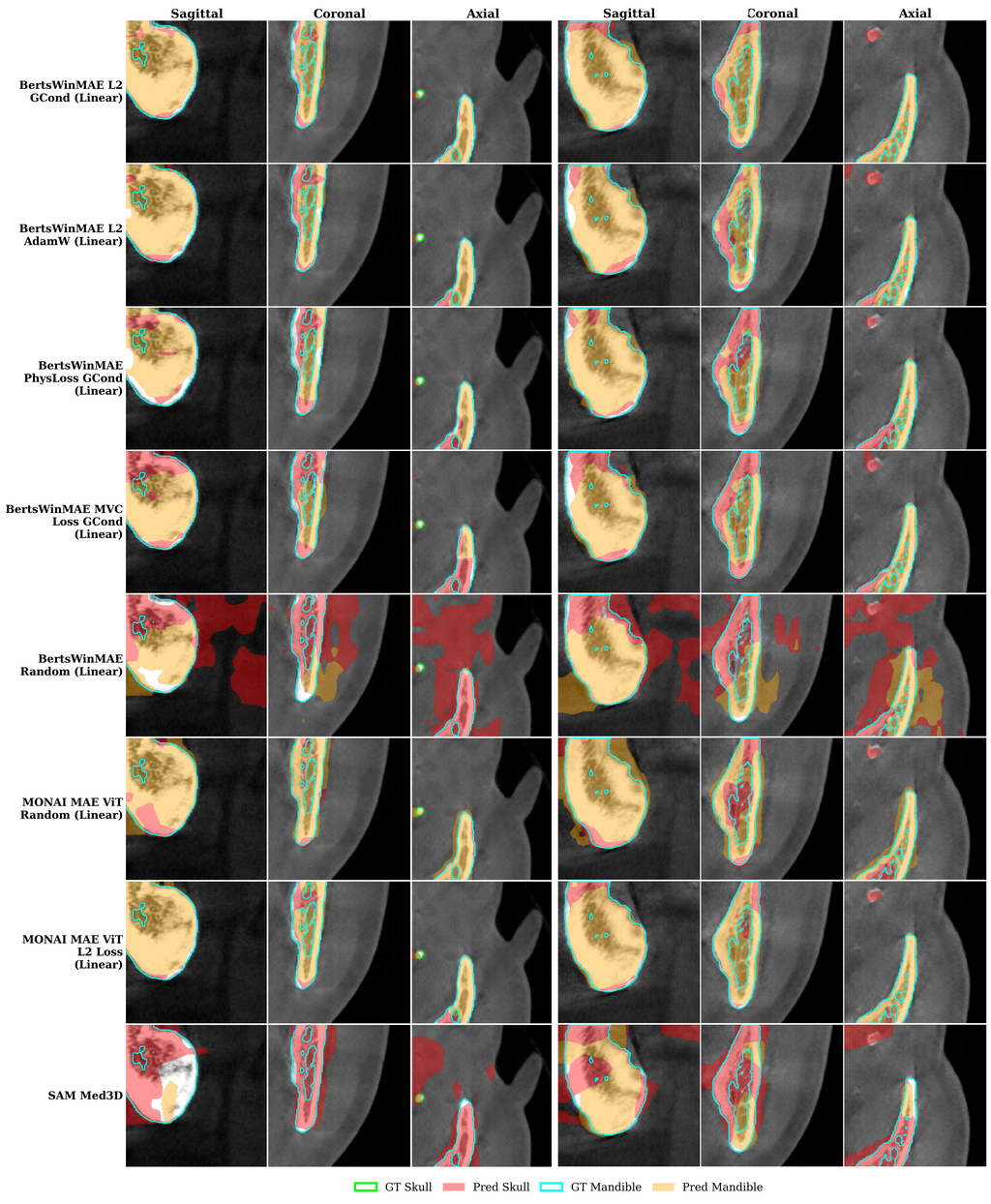}
    \caption{Qualitative comparison of multi-class TMJ segmentation on challenging cases. The results for two representative temporomandibular joints (Right and Left panels) across three orthogonal anatomical planes: Sagittal, Coronal, and Axial. Rows correspond to different self-supervised encoders evaluated via linear probing. Ground Truth (GT) annotations are depicted as contours, while model predictions are overlaid as semi-transparent masks.}
    \label{fig:qual_seg}
\end{figure}

Qualitative error analysis was conducted on the most challenging validation subject, identified by the lowest consistent Dice scores across all models (Fig.~\ref{fig:qual_seg}). Visual inspection revealed specific interaction patterns driven by acquisition anomalies. First, operator-induced caudal shifts resulted in the systematic misclassification of the C1 transverse process as mandibular bone across all linear probes, indicating a reliance on local texture statistics over global positional encoding. Second, regarding label noise, pre-trained encoders consistently classified ossified stylohyoid ligaments as background, diverging from the automated Ground Truth which labeled these ambiguous structures as bone. This suggests the unsupervised acquisition of a robust global shape prior capable of filtering high-frequency annotation artifacts.
\vfill 
\newpage 
\section{DISCUSSION}

\subsection{Training dynamics and convergence analysis}
Our analysis identifies a critical limitation in current 3D self-supervised paradigms: \textit{topological sparsity}. We demonstrate that while the standard asymmetric MAE approach offers per-iteration computational efficiency, the random removal of 75\% of tokens disrupts the global geometric priors essential for volumetric reconstruction. By reintroducing a \textbf{Full Token Grid}, BertsWin preserves a continuous spatial manifold, enabling the Swin Transformer's shifted window mechanism to operate on a topologically complete context. This architectural shift fundamentally alters the optimization landscape, as evidenced by the contrasting convergence trajectories (Fig.~\ref{fig:convergence}). While the sparse MAE ViT baseline exhibits a prolonged ``warm-up'' phase and transient instability (steps 25k--75k) likely driven by the difficulty of resolving 3D coherence from fragmented inputs, BertsWin demonstrates a monotonic, stable reduction in error from the earliest iterations.

To disentangle architectural contributions from optimization effects, we systematically decoupled the sources of acceleration. Under identical optimization conditions (AdamW), the transition from sparse masking to full-grid processing yielded a $5.8$-fold speedup in convergence (660 vs 114 epochs). When coupled with the GCond optimizer, the framework achieved a cumulative $15.0$-fold reduction in training duration required to reach SOTA reconstruction fidelity. These results challenge the prevailing assumption that sparse masking is a prerequisite for 3D efficiency. At standard resolutions ($224^3$, patch size $16^3$), the linear complexity of the Swin Transformer effectively neutralizes the overhead of processing masked tokens, converting the ``topological completeness'' into a net computational gain rather than a throughput trade-off.

\begin{figure}[H]
    \centering
    \begin{subfigure}[b]{0.49\linewidth}
        \centering
        \includegraphics[width=\linewidth]{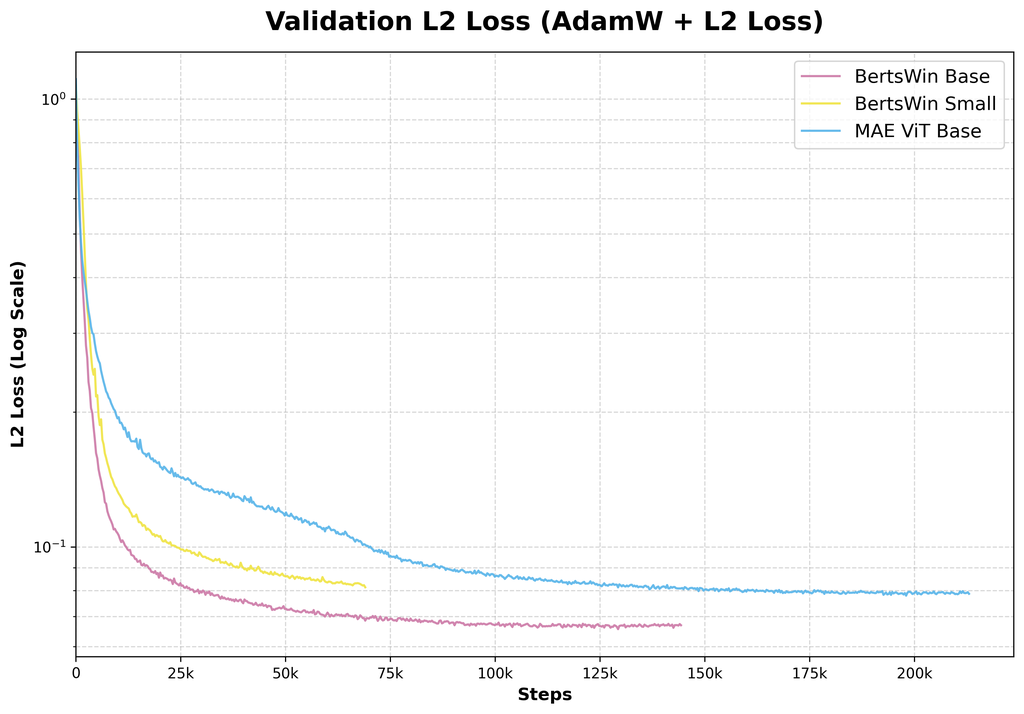}
        \caption{L2 Loss}
    \end{subfigure}
    \begin{subfigure}[b]{0.49\linewidth}
        \centering
        \includegraphics[width=\linewidth]{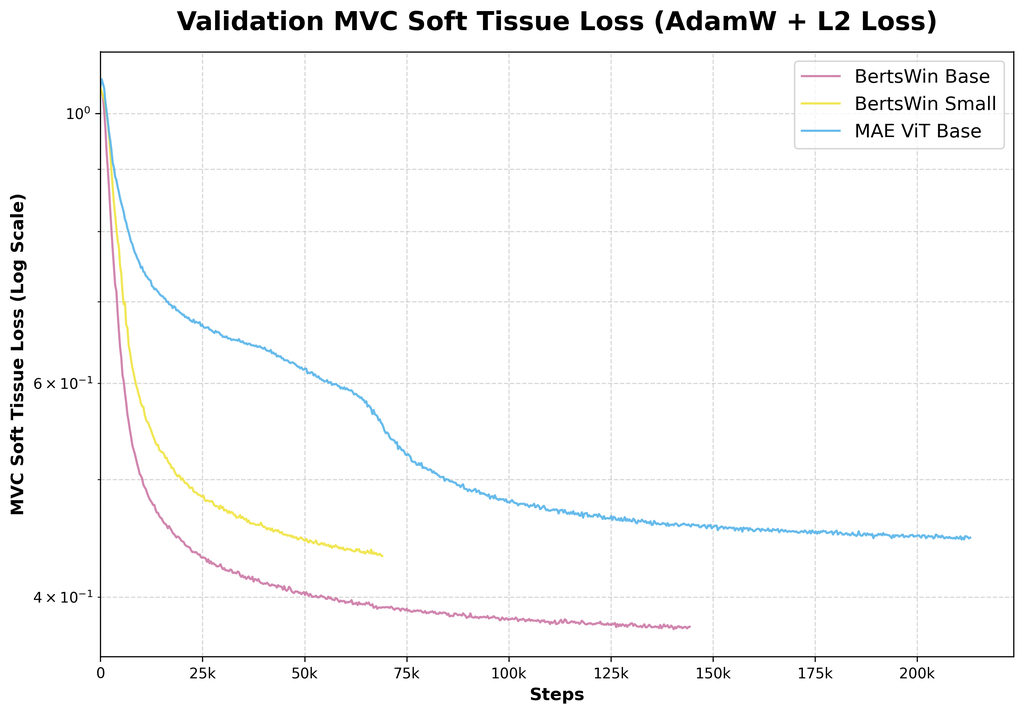}
        \caption{Soft Tissue Error}
    \end{subfigure}
    \caption{Decoupling Architectural Efficiency from Model Capacity: Ablation Study on Convergence Dynamics. Validation trajectories for L2 Loss (Left) and Soft Tissue Structural Error (Right) under identical optimization (AdamW + L2).}
    \label{fig:ablation_conv}
\end{figure}

To rigorously rule out model capacity as a confounding factor, we conducted an ablation study comparing a lightweight configuration, BertsWin-Small ($\approx 25\%$ parameters, early termination at 69k steps), against the fully converged ViT-Base baseline (198k steps). Remarkably, despite the significant deficit in capacity and training time, BertsWin-Small surpassed the SOTA baseline in soft tissue structural fidelity ($1-\text{MVC}$: $0.432$ vs $0.446$) and achieved parity in global L2 reconstruction (Fig.~\ref{fig:ablation_conv}). This confirms that the observed performance advantage is intrinsic to the architectural design -- specifically the preservation of the 3D grid topology -- rather than parameter count or extended training regimes. Consequently, we posit that preserving the complete volumetric topology is a prerequisite for efficient semantic convergence in medical imaging.

\subsection{Comparative analysis of latent space representations}

A comparison of the density of latent representations (Fig.~\ref{fig:latent_density}) showed that the GCond guided frameworks maintain a broad dynamic range with a structured shift in activation density, demonstrating that the model learns a non-trivial, semantically rich manifold without succumbing to the representation collapse typical of voxel-wise MSE optimization.

Analysis of intrinsic dimensionality using \textit{Effective Rank} ($R_{\text{eff}}$) of the feature covariance matrix (Table~\ref{tab:latent_topology}) revealed critical divergence between pixel-based baselines and structural encoders. The MAE ViT baseline exhibits a high effective rank ($R_{\text{eff}} \approx 23.36$). However, coupled with its low geometric invariance ($0.549$), this high dimensionality suggests overfitting to high-frequency spectral noise inherent in CBCT, rather than semantic richness. In contrast, the proposed BertsWin (PhysLoss GCond) compresses the representation to a low effective rank ($R_{\text{eff}} \approx 4.01$). Crucially, this aligns closely with the zero-shot SAM Med3D foundation model ($R_{\text{eff}} \approx 4.54$), while significantly diverging from the noise-dominated baseline. We interpret this convergence as the isolation of fundamental geometric primitives (e.g., planar surfaces, curvature gradients) common to anatomical structures. Instead of memorizing ROI-specific textures (which would inflate the rank), the Physics-Informed loss forces the encoder to learn a parsimonious manifold of shape descriptors. This compact representation suggests the model captures generalized anatomical coherency rather than specific TMJ texture patterns.

A critical finding is that loss function selection dominates latent space topology, imposing a performance ceiling that advanced optimizers cannot overcome. All models trained with mean squared error including MAE ViT, BertsWin with AdamW, and BertsWin with GCond converged to suboptimal local minima characterized by low geometric invariance (median $0.549\text{--}0.628$) and poor intra-patient similarity (median $< 0.53$) (Table~\ref{tab:latent_topology}). Significantly, applying GCond to the L2 loss did not yield significant improvements over AdamW. This implies that the MSE loss surface itself lacks the necessary curvature to guide the optimizer toward anatomically meaningful minima. By averaging errors pixel-wise, MSE acts as a low-pass filter, erasing the high-frequency details required for unique patient identification.

Transitioning to structural objectives (MVC and PhysLoss) resulted in a pronounced quantitative shift in representation quality. The median Intra-Patient similarity increased significantly from $0.472$ (L2) to $0.791$ (PhysLoss) (Table~\ref{tab:latent_topology}). Given that encoders were trained exclusively on right-sided joints, validation on contralateral left joints constitutes an out-of-distribution test. The poor performance of L2-based models implies overfitting to the specific intensity distribution of the right-sided training set. Conversely, the high similarity score of PhysLoss confirms that the Physics-Informed objective forces the encoder to learn chirality-invariant structural features that remain robust even under anatomical mirroring.

While the initialized BertsWin (Random) exhibits a deceptively competitive mean geometric invariance ($0.645$), statistical dispersion analysis reveals its lack of semantic stability. The IQR for the random network is $0.476$ -- substantially wider than the optimized BertsWin (PhysLoss) ($\text{IQR} = 0.140$). Furthermore, the random baseline fails to segregate patients, showing a high positive Inter-Patient similarity (median = $0.280$), whereas the optimized model effectively orthogonalizes distinct patients (median = $-0.120$). Thus, the high mean invariance in random networks is attributable to the preservation of local low-level topology via random projection, lacking the discriminative semantic structure achieved through training.

\subsection{Sensitivity analysis: anatomical chirality and orientation}

A paired Wilcoxon signed-rank test comparing Intra-Patient similarity (Real Right vs Mirrored Contralateral Left) against a Symmetry Check (Real Right vs Artificially Mirrored Right) showed that BertsWin (PhysLoss GCond) demonstrates a statistically significant preference for the Intra-Patient pairs ($p < 0.001$), with a median similarity differential of $-0.039$ favoring the contralateral joint over the synthetic reflection. This indicates that the PhysLoss objective enforces a strictly oriented structural representation. The encoder is sensitive to the biological directionality of the bone structure: it recognizes that a ``flipped left'' joint conforms to the learned rules of right-sided anatomy, whereas a ``flipped right'' joint violates them. 

In contrast, the L2-based models (e.g., BertsWin L2 GCond) show negligible difference (median differential $\approx 0$, $p = 0.72$), implying a lack of sensitivity to anatomical orientation. This suggests L2 optimization leads to ``isotropic'' feature extractors that rely on local texture statistics -- which remain constant under mirroring -- rather than global shape configuration. Thus, the PhysLoss model's rejection of synthetic symmetry violations validates its semantic specificity to valid anatomical configurations rather than representing failure of invariance.

At the same time, violin plots visualizing the density distribution of cosine similarity scores across three evaluation protocols showed that the BertsWin (PhysLoss) demonstrates the largest discriminative margin (separation) between inter-patient noise and intra-patient signal, significantly outperforming L2-based baselines which exhibit overlapping distributions (Fig.~\ref{fig:violins}).

While scatter plots comparing feature similarity for real contralateral pairs versus artificially mirrored pairs (Fig.~\ref{fig:scatter}) showed that the BertsWin (PhysLoss) model produces features that are much more consistent for morphologically correct pairs (high similarity, top-right) compared to baseline models, which fail to learn features with high confidence (low similarity, bottom-left). This confirms that the PhysLoss objective enforces strict sensitivity to anatomical orientation without sacrificing the global structural coherence required for high-confidence feature extraction.

\subsection{Reconstruction fidelity and loss function priors}

Qualitative analysis of orthogonal projections (Fig.~\ref{fig:2d_slice}) and volumetric isosurfaces (Fig.~\ref{fig:3d_iso}) confirms that all evaluated models successfully recover the global mandibular geometry. However, closer inspection reveals fundamental disparities in textural fidelity and spatial coherence driven by the choice of optimization objective.

Models trained with standard L2 loss (both BertsWin-L2 and MONAI MAE ViT) exhibit characteristic over-smoothing artifacts. In regions defined by stochastic texture -- such as soft tissue fascia and trabecular bone -- these models effectively function as low-pass filters, suppressing high-frequency gradients in favor of mean intensity values. While BertsWin-L2 achieves optimal spatial coherence, yielding monolithic volumes where patch boundaries are seamless, the resulting surfaces suffer from a ``plastic effect,'' reflecting a systematic loss of cortical micro-relief.

In contrast, the MONAI MAE ViT baseline displays severe fragmentation, with distinct patch boundaries creating blocking artifacts (Fig.~\ref{fig:3d_iso}). This indicates a failure to establish global volumetric context, likely due to the model's inability to bridge the aggressive 75\% masking gap without the inductive bias of a continuous grid.

The proposed BertsWin configurations guided by structural objectives (PhysLoss/MVC with GCond) occupy a distinct functional niche. They yield significantly sharper reconstructions with superior preservation of internal trabecular structure and cortical boundaries. However, this spectral fidelity comes with a trade-off: these models exhibit minor residual traces at patch boundaries -- though significantly less pronounced than the MONAI baseline. We attribute this to the asynchronous convergence of the multi-component loss: the optimization of high-frequency local texture (correlation/gradient terms) competes with the minimization of global continuity errors, requiring more iterations to resolve boundary artifacts than permitted by standard early stopping criteria.

Finally, while structural losses significantly enhance realism, the recovered micro-texture remains less distinct than the ground truth. This limitation aligns with findings by Oh J. et al. \cite{Oh2024}, attributing the bottleneck to the capacity constraints of lightweight decoders, which struggle to propagate the full high-frequency code from the latent space to the voxel grid. Nevertheless, BertsWin successfully balances global topology with local structural integrity, substantially surpassing SOTA baselines in diagnostic plausibility.

\subsection{Downstream segmentation performance}

\begin{table*}[t]
    \centering
    \caption{Comparison of Segmentation Performance with a 2-Layer Non-Linear Decoder. Values represent mean $\pm$ standard error.}
    \label{tab:nonlinear_seg}
    \small
    \begin{tabular}{lcccccc}
        \toprule
        \multirow{2}{*}{\textbf{Model}} & \multicolumn{3}{c}{\textbf{Mandible}} & \multicolumn{3}{c}{\textbf{Skull}} \\
        \cmidrule(lr){2-4} \cmidrule(lr){5-7}
        & \textbf{Dice} & \textbf{HD95} (mm) & \textbf{ASSD} (mm) & \textbf{Dice} & \textbf{HD95} (mm) & \textbf{ASSD} (mm) \\
        \midrule
        BertsWinMAE L2 GCond & \textbf{0.822 $\pm$ 0.004} & 2.24 $\pm$ 0.05 & 0.56 $\pm$ 0.05 & 0.935 $\pm$ 0.002 & 1.38 $\pm$ 0.05 & 0.24 $\pm$ 0.00 \\
        BertsWinMAE PhysLoss GCond & 0.811 $\pm$ 0.001 & 2.45 $\pm$ 0.07 & 0.69 $\pm$ 0.04 & 0.929 $\pm$ 0.003 & 1.50 $\pm$ 0.08 & 0.28 $\pm$ 0.01 \\
        MONAI MAE ViT L2 Loss & \textbf{0.822 $\pm$ 0.002} & \textbf{2.21 $\pm$ 0.06} & \textbf{0.50 $\pm$ 0.03} & \textbf{0.937 $\pm$ 0.001} & \textbf{1.28 $\pm$ 0.05} & \textbf{0.20 $\pm$ 0.02} \\
        SAM Med3D & 0.702 $\pm$ 0.000 & 3.61 $\pm$ 0.04 & 0.99 $\pm$ 0.03 & 0.860 $\pm$ 0.002 & 2.07 $\pm$ 0.04 & 0.52 $\pm$ 0.02 \\
        \bottomrule
    \end{tabular}
\end{table*}

A critical observation emerged from the initial linear probing analysis (Table~\ref{tab:segmentation}), where models trained with structural objectives (PhysLoss, MVC) paradoxically exhibited higher Hausdorff Distances (HD95) compared to L2-optimized baselines. We hypothesized that this metric discrepancy did not reflect a degradation of feature quality, but rather the intrinsic limitations of a linear projection head in resolving the complex, non-linear boundary semantics encoded by structural losses. The application of a lightweight non-linear decoder validated this hypothesis, bridging the performance gap and revealing the true semantic capacity of the representation. 

As detailed in Table~\ref{tab:nonlinear_seg}, the HD95 for PhysLoss models on the Skull segmentation task decreased dramatically from $10.04$ mm (linear probe) to $1.50$ mm (non-linear decoder) --- a reduction of over 75\%. While L2-optimized models (MONAI, BertsWin-L2) retained a slight metric advantage in the linear regime due to the inherent alignment between MSE optimization and the density-thresholded Ground Truth ($\text{HD95} \approx 2.2$ mm), the structural models demonstrated superior recovery of complex topology when decoded non-linearly. This confirms that high-fidelity boundary information is fully encoded in the latent space of BertsWin, albeit within a non-linear manifold that requires a simplified decoding layer to map onto discrete density-based annotations.

Spectral analysis elucidates the mechanism behind this behavior. The MAE ViT baseline exhibited a high Effective Rank ($R_{\text{eff}} \approx 23.36$) coupled with low geometric invariance (Table~\ref{tab:latent_topology}). In this context, a high $R_{\text{eff}}$ indicates overfitting to high-frequency spectral noise --- the ``Cone Effect'' --- rather than feature richness. The associated low geometric invariance confirms that this feature space is primarily occupied by non-discriminative, high-entropy texture features which the pixel-wise L2 loss failed to regularize. In contrast, BertsWin guided by PhysLoss compresses the representation to a significantly lower effective rank ($R_{\text{eff}} \approx 4.01$). This compression signals the isolation of fundamental geometric primitives rather than texture memorization. While such a compact, low-rank manifold presents a challenge for linear separation (which favors high-dimensional sparsity), it contains superior structural information, as evidenced by the robust performance of the non-linear decoder.

Our comparative analysis with the SAM Med3D foundation model established critical limits for cross-domain generalization in high-precision osteology. Despite utilizing a robust generalist encoder, SAM Med3D performed significantly worse in the zero-shot linear probing regime ($\text{DSC} < 0.70$). Even when afforded a non-linear decoder (Table~\ref{tab:nonlinear_seg}), the foundation model failed to achieve parity with domain-specific encoders (DSC $0.702$ vs $0.822$). This performance disparity highlights a substantial ``spectral domain gap'': generalist models trained primarily on soft-tissue dominant imaging (MRI/CT) lack the specific mineralized density priors inherent to dental CBCT. Furthermore, generalist encoders failed to capture anatomical chirality --- the ability to distinguish Left/Right anatomical features --- which the specialized BertsWin model acquired spontaneously. This observation validates the necessity of domain-specific pre-training to enforce ``biological directionality'' constraints that generic foundational models do not possess.

The choice of optimizer played a decisive role in boundary definition. The GradientConductor (GCond) optimizer consistently improved precision metrics compared to AdamW, reducing the mean surface error (ASSD) by approximately 23\% for the Mandible. This validation confirms GCond's role in promoting more robust latent boundaries, independent of the loss function. Furthermore, the robust performance of the BertsWinMAE (L2 GCond) configuration, which matches SOTA baselines in the linear probing setup, confirms that the hybrid Swin-based architecture successfully learns spatial topology comparable to heavier, computationally intensive ViT models.

To investigate the drivers of metric degradation in specific samples, we conducted a rigorous visual inspection of failure cases (Fig.~\ref{fig:qual_seg}). This analysis revealed that the divergence between model predictions and pseudo-Ground Truth is often driven by semantic disagreements rather than reconstruction failure.

\noindent \textbf{1. Osseous Boundary Definition.} Reference masks typically adhere to high-density cortical layers, resulting in ``lacunae'' (gaps) within trabecular regions. L2-optimized models minimize intensity variance and thus converge to these local density statistics, effectively replicating the lacunae and achieving deceptively low HD values. Conversely, the PhysLoss objective promotes volumetric closure, generating topologically continuous anatomical volumes that systematically deviate from the discretized, density-derived reference map. Consequently, the elevated HD95 in structural models often reflects a penalty for topological completeness rather than structural hallucination.

\noindent \textbf{2. Robustness to Label Noise.} In regions with anatomical ambiguity, such as ossified stylohyoid ligaments labeled as mandible by the automated pipeline, pre-trained encoders consistently classified these structures as background. This indicates that self-supervised pre-training establishes a robust global shape prior that mitigates annotation noise, prioritizing learned anatomical consistency over high-frequency label artifacts.

\noindent \textbf{3. Positional Feature Conflict.} Positional conflicts caused by operator-induced shifts (e.g., misclassification of the C1 vertebra) suggest that the linear probe relies primarily on local bone texture. However, the architectural comparison using random initialization confirms that the BertsWin stem structure requires learned weights to propagate signals effectively, whereas convolutional baselines (MONAI) retain some inductive bias even without training. Thus, the superior performance of the pre-trained BertsWin confirms that domain-specific SSL yields feature spaces with significantly improved linear separability and semantic integrity.

\subsection{Context-driven volumetric inpainting (anatomical hallucination)}

To evaluate semantic integrity, we implemented a \textit{Contextual Shell Masking} protocol designed to test generative capacity under pre-training conditions (75\% masking). By iteratively expanding the mandibular mask into surrounding tissues until only 25\% of the volume remained visible, we forced the model to reconstruct complex geometry relying solely on the learned biomechanical logic of the craniofacial complex, thereby minimizing out-of-distribution shifts. Reconstructions were evaluated using a fixed 400 HU threshold to decouple mineralized structure from soft tissue artifacts.
\begin{figure}[H]
    \centering
    \includegraphics[width=0.45\textwidth]{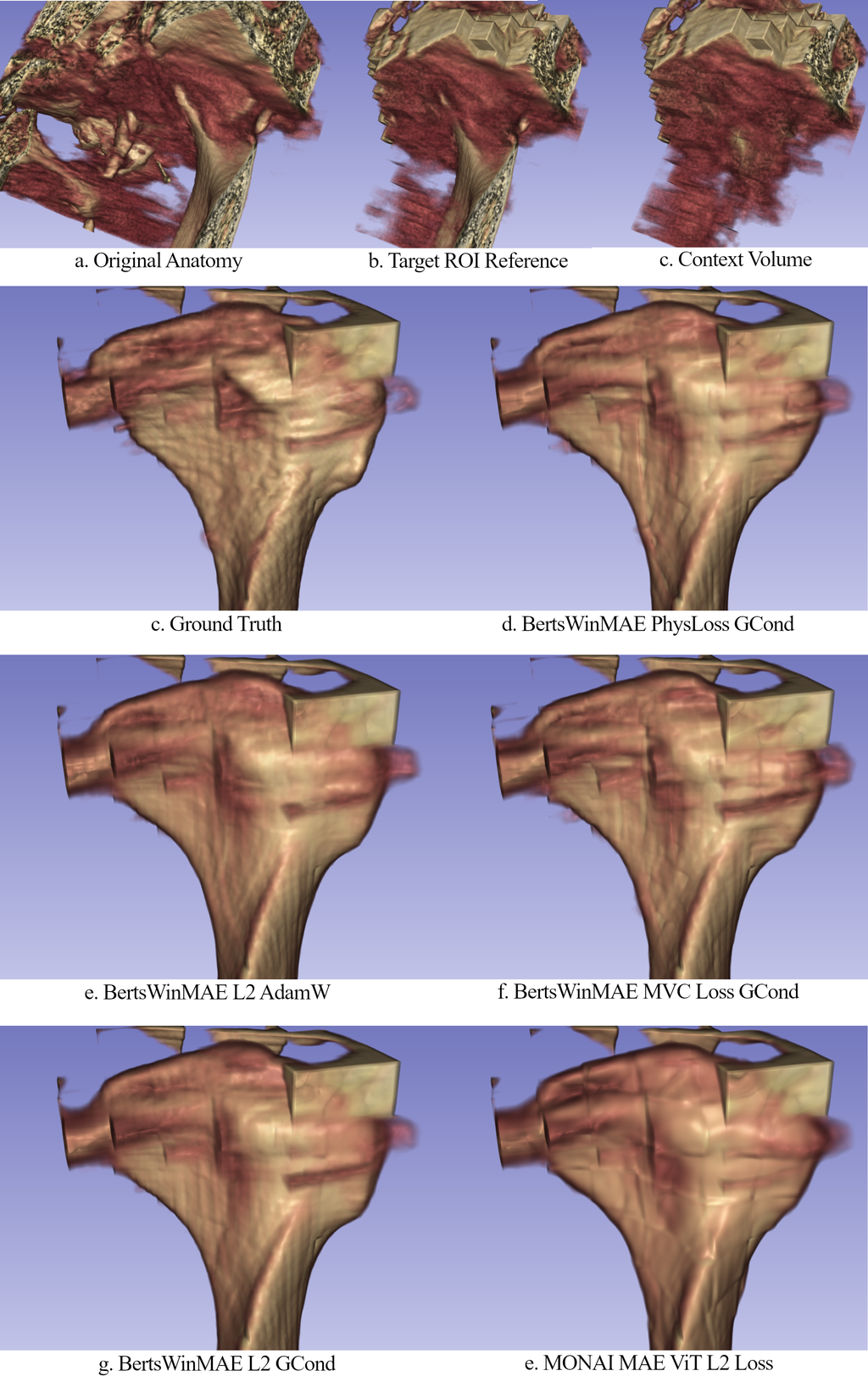}
    \caption{Context-Driven Volumetric Inpainting under controlled 25\% visibility. Note: Isosurface renderings of the mandibular reconstruction task. (a) Original Anatomy; (b) Target ROI Reference; (c) Input Context; (d) Ground Truth.}
    \label{fig:shell_mask}
\end{figure}
Quantitative analysis reveals a fundamental trade-off driven by the optimization objective. BertsWin (L2 GCond) maximized volumetric overlap (Dice $0.955$, ASSD $0.11$ mm), confirming that MSE establishes a robust mean shape prior. Conversely, BertsWin (MVC) demonstrated superior boundary precision with an HD95 of $0.49$ mm, significantly outperforming L2-optimized models ($\ge 0.80$ mm). This indicates that structural loss constrains localized deviations, tightening the geometric envelope. The MONAI baseline yielded the lowest fidelity (Dice $0.941$) and highest intensity error (RMSE $297.2$ HU), reflecting reduced capacity to match the target distribution.

Qualitative inspection (Fig.~\ref{fig:shell_mask}) corroborates these metrics. L2-optimized models exhibit spectral smoothing (``plastic effect''), where trabecular heterogeneity is averaged into uniform density fields. In contrast, PhysLoss and MVC configurations recover high-frequency textures, though MVC displays minor grid artifacts attributable to the asynchronous convergence of structural loss components. Notably, the faithful reconstruction of complex osseous geometry from a spatially restricted context suggests the encoder captures invariant tissue physics rather than memorizing rigid organ-specific topology. We hypothesize this reflects a ``holographic'' redundancy, where local density gradients within the limited mandibular fragments contain sufficient inductive bias to extrapolate the integral anatomical structure.

\begin{figure}[h]
    \centering
    \includegraphics[width=0.45\textwidth]{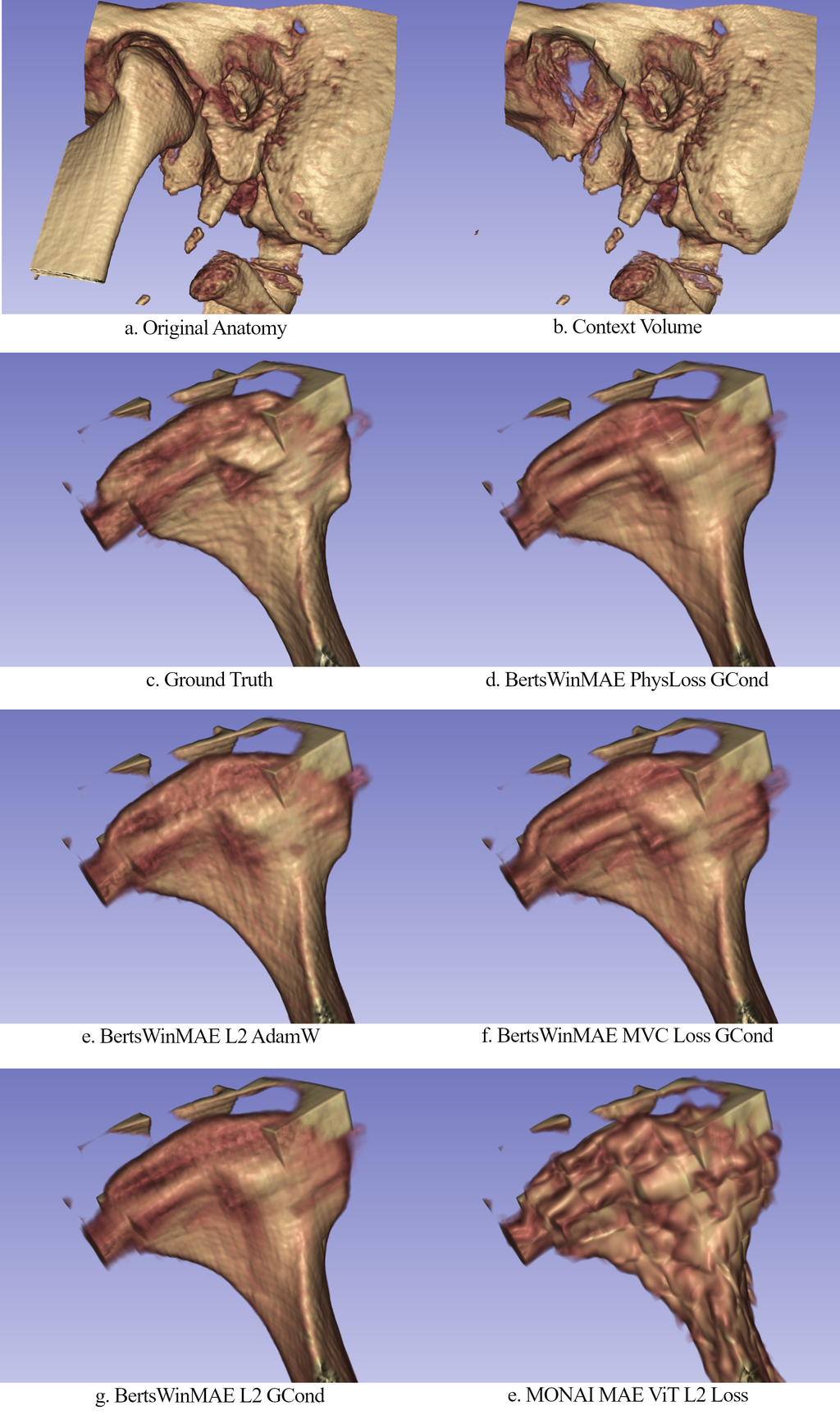}
    \caption{3D volumetric reconstruction of the mandible Out-of-Distribution(OOD) Regime. Note: The isosurface renderings were generated in 3D Slicer with soft tissues removed to visualize osseous structures. (a) Original Anatomy; (b) Input context (93\% visible); (c) Ground Truth.}
    \label{fig:ood_stress}
\end{figure}

\begin{figure*}[t]
    \centering
    \includegraphics[width=\textwidth]{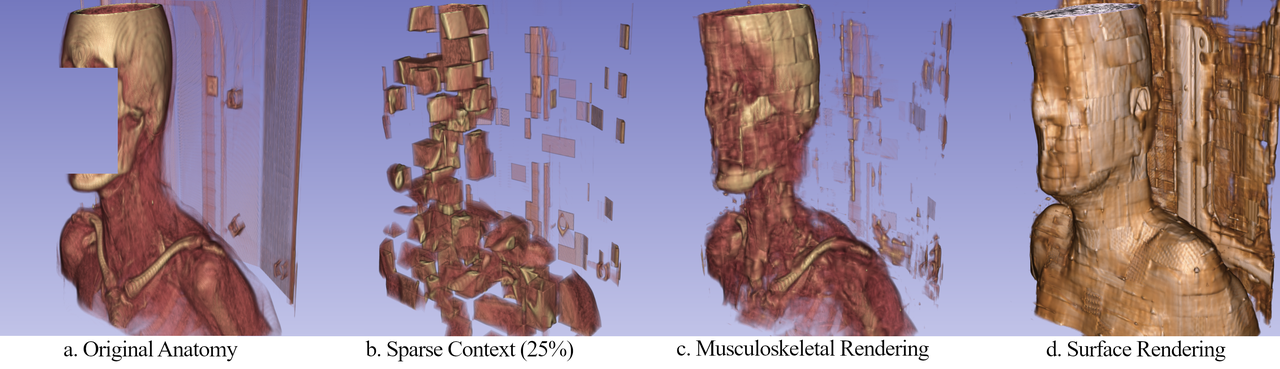}
    \caption{Emergent Generalization and Holographic Reconstruction of Unseen Anatomy. Note: (a-b) Ground Truth vs 25\% sparse input of the head and upper torso. (c) Volumetric inference (Bone/Muscle). Despite training exclusively on localized TMJ crops, the model spontaneously reconstructs out-of-distribution structures, including the clavicle and cervical spine. (d) Soft tissue rendering. The network extrapolates coherent global facial geometry (nose, ear) while abstracting high-frequency features (eyes/mouth), supporting the hypothesis that the encoder learns invariant local tissue physics rather than semantic organ topology.}
    \label{fig:radcure}
\end{figure*}

To validate the stability of these learned representations, we subjected the encoders to an Out-of-Distribution (OOD) ``stress test'' with 93\% visible context, effectively inverting the training masking ratio (25\%). Under this high-density regime, the baseline MONAI ViT exhibited severe feature collapse, characterized by ``globular'' artifacts and massive volume overestimation (111.2\% of GT). This output instability indicates that the deterministic decoder succumbed to attention dilution, where global self-attention mechanisms fail to generalize to token densities significantly exceeding the training distribution.

In stark contrast, BertsWin maintained high topological fidelity across all configurations (Dice $> 0.90$), validating the architectural advantage of the Swin Transformer. The local shifted-window mechanism effectively enforces signal consistency, preventing the ``feature poisoning'' observed in global ViTs. Quantitatively, the PhysLoss-guided model demonstrated the highest robustness, achieving near-perfect volumetric alignment (99.1\%) and correctly reconstructing delicate anatomical features such as the condylar process (HD95 $0.57$ mm). While the L2-optimized BertsWin models maintained geometric topology, they exhibited distinct biases: the L2-AdamW configuration preserved fine details, whereas the L2-GCond configuration suffered from volumetric underestimation (85.7\%). These results confirm that while the BertsWin architecture provides intrinsic immunity to context density shifts, the Physics-Informed loss is essential for calibrating the volumetric scale.

\subsection{Emergent Generalization and the Holographic Nature of Anatomical Priors}

The model's ability to generalize from a restricted training domain (local CBCT patches of the TMJ) to entirely unseen anatomical regions (Fig.~\ref{fig:radcure}) (e.g., thoracic cage, cervical spine) and imaging modalities (standard CT with varying voxel spacing) raises fundamental questions regarding the nature of representations learned via masked autoencoding. The successful volumetric reconstruction of ribs and soft facial tissues -- structures absent from the training distribution -- suggests that the encoder has transcended the memorization of organ-specific topology.

We hypothesize that by forcing the model to solve highly ambiguous reconstruction tasks (75\% masking) within a high-resolution local context, the network approximates the fundamental stochastic rules of biological morphogenesis rather than high-level semantic labels. The learned dependencies such as cortical continuity, density gradients at tissue interfaces, and trabecular texture coherence -- appear to be invariant across anatomical scales and locations. This finding implies a degree of local self-similarity in volumetric medical data, where the ``physics'' of tissue organization learned on a micro-scale contains sufficient inductive bias to reconstruct macro-scale anatomy. This suggests a paradigm shift where robust foundation models could potentially be derived from high-fidelity local datasets by learning the universal ``grammar'' of biological matter, rather than requiring exhaustive atlas-level coverage.

\subsection{Current Limitations}
Several limitations of this study warrant consideration. First, our evaluation focused primarily on TMJ imaging using CBCT data. While the out-of-distribution experiments suggest broader applicability, systematic evaluation across diverse anatomical regions and imaging modalities such as MRI, ultrasound, PET is needed to establish generalizability. 

Second, the anatomically weighted loss function requires semantic segmentation masks for bone surfaces and soft tissues, which may not be readily available for all anatomical regions or may require additional annotation effort. 

Third, given the linear complexity of the BertsWin encoder at high resolutions ($512^3$, P16), the training bottleneck shifts from GPU compute to Data I/O bandwidth. While the architecture supports efficient high-resolution processing, current PCIe bandwidth limits may constrain throughput during full-grid training.

Finally, to rigorously quantify convergence acceleration, we employed an early stopping criterion for BertsWin that resulted in significantly lower total sample exposure compared to the fully converged baselines. While this successfully demonstrated the architecture's topological efficiency, the reported performance metrics likely represent a conservative lower bound of the model's capacity. Future work exploring extended training regimes could potentially yield further refinements in the latent feature space, albeit at the cost of the computational efficiency gains highlighted in this study.

\section{CONCLUSION}
This study presents BertsWin, a hybrid self-supervised framework that resolves the critical conflict between computational efficiency and topological integrity in 3D medical imaging. Our findings demonstrate that preserving a complete volumetric manifold via a Full Token Grid is a prerequisite for efficient semantic convergence. By enabling local attention mechanisms to operate on a topologically continuous context, BertsWin achieves a $15$-fold reduction in optimization steps compared to sparse MAE baselines. Crucially, we prove that this acceleration secures a net reduction in total computational resources, challenging the prevailing assumption that aggressive token masking is required for 3D efficiency.

Furthermore, we show that transitioning from generic pixel-wise objectives to Anatomically Informed (PhysLoss) functions significantly enhances representation quality, mitigating the ``feature collapse'' typical of MSE optimization. The successful generative reconstruction of complex osseous geometry from limited context confirms that the encoder captures invariant local tissue physics -- the ``grammar of biological matter'' -- rather than merely memorizing organ-specific topology. In conclusion, BertsWin establishes a principled paradigm for volumetric representation learning, validating that architectural coherence and domain-specific objectives are decisive factors in achieving SOTA performance for specialized medical tasks.

\section{DATA AVAILABILITY}
The source code and experimental data supporting this study are publicly available in a Zenodo repository (DOI: 10.5281/zenodo.17916932) and on GitHub at the following repositories:
\begin{itemize}
    \item GCond: \url{https://github.com/AlevLab-dev/GCond}
    \item BertsWinMAE: \url{https://github.com/AlevLab-dev/BertsWinMAE}
\end{itemize}

\end{document}